\documentclass{article}

\usepackage[preprint,nonatbib]{neurips_2024}

\usepackage[utf8]{inputenc} % allow utf-8 input
\usepackage[T1]{fontenc}    % use 8-bit T1 fonts
\usepackage{hyperref}       % hyperlinks
\usepackage{url}            % simple URL typesetting
\usepackage{booktabs}       % professional-quality tables
\usepackage{amsfonts}       % blackboard math symbols
\usepackage{nicefrac}       % compact symbols for 1/2, etc.
\usepackage{microtype}      % microtypography
\usepackage{xcolor}         % colors
\usepackage[backend=biber, style=numeric, sorting=none, maxnames=10]{biblatex}

\usepackage{tikz}
\usetikzlibrary{shapes,arrows,positioning}
\usepackage{subcaption}
\usepackage{multirow} % I added
\usepackage{algorithm} % I added
\usepackage{algpseudocode} % I added
\usepackage{amsmath} % I added
\usepackage{amssymb} % I added
\usepackage{adjustbox} % I added

\title{MedConceptsQA: Open Source Medical Concepts QA Benchmark}

\author{%
  Ofir Ben Shoham \\
  Department of Software and Information Systems Engineering\\
  Ben-Gurion University of the Negev\\
  \texttt{benshoho@post.bgu.ac.il} \\
  \And
  Nadav Rappoport \\
  Department of Software and Information Systems Engineering\\
  Ben-Gurion University of the Negev\\
\texttt{nadavrap@bgu.ac.il} \\
}
\begin{document}

\maketitle

% Abstract
\begin{abstract}
We present MedConceptsQA, a dedicated open source benchmark for medical concepts question answering. The benchmark comprises of questions of various medical concepts across different vocabularies: diagnoses, procedures, and drugs. The questions are categorized into three levels of difficulty: easy, medium, and hard. We conducted evaluations of the benchmark using various Large Language Models. Our findings show that pre-trained clinical Large Language Models achieved accuracy levels close to random guessing on this benchmark, despite being pre-trained on medical data. However, GPT-4 achieves an absolute average improvement of nearly 27\%-37\% (27\% for zero-shot learning and 37\% for few-shot learning) when compared to clinical Large Language Models. Our benchmark serves as a valuable resource for evaluating the understanding and reasoning of medical concepts by Large Language Models. Our benchmark is available at \href{https://huggingface.co/datasets/ofir408/MedConceptsQA}{https://huggingface.co/datasets/ofir408/MedConceptsQA}
\end{abstract}

% Introduction and related work.
\section{Introduction}

% Large Language Models background 
Large Language Models (LLMs) are trained on huge datasets and comprising billions of parameters. LLMs have demonstrated remarkable effectiveness across a range of language-related tasks, including question-answering and text generation \cite{zhao2023survey}. Their applicability extends beyond traditional language domains to areas such as robotics, cybersecurity, law, and also medicine \cite{kaddour2023challenges}.

% Clinical Large Language Models (the models) + downstream tasks of clinical llm. prediction/diagnosis/summarization. 
Clinical LLMs (CLLMs) are LLMs that are trained on medical datasets. To name a few: Meditron \cite{chen2023meditron}, BioMistral \cite{labrak2024biomistral}, BioMedGPT \cite{luo2023biomedgpt}, GatorTron \cite{yang2022large}, and Meerkat \cite{kim2024small}. CLLMs have shown their effectiveness to be used in tasks such as clinical text classification \cite{shoham2023cpllm}, medical chatbots, healthcare education, and clinical text generation \cite{he2024foundation}. % (for example, report generation).

Clinical data is frequently represented using standardized medical codes rather than natural language descriptions. Therefore, it's essential that CLLMs understand the meaning of these medical codes and their differences. One way to test the understanding of CLLMs is to evaluate their ability to interpret clinical codes.

% Similar clinical benchmarks & QA benchmarks and differences. 
Existing clinical benchmarks are available for evaluating CLLMs. For example, BioASQ-QA \cite{krithara2023bioasq} is a manually curated corpus of Biomedical Question Answering (QA) for documented retrieval, text snippets extraction, and summarization. EmrQA \cite{pampari2018emrqa} is a corpus for QA on Electronic Medical Records based on clinical notes. PubMedQA \cite{jin-etal-2019-pubmedqa} is a dataset that contains biomedical QA based on PubMed abstracts. In addition to QA datasets, there are clinical datasets in other fields such as classification \cite{ojo2023medai} and summarization \cite{molla2016corpus, dada2024clue}. Soroush et al. \cite{soroush2024large} used real clinical visit data to provide LLMs with a code description and prompted them to generate a billing code. They found that all general-purpose LLMs tested performed poorly at this task. Notably, they did not evaluate CLLMs that were trained on medical data and they only used zero-shot. We used a different approach for evaluation: our proposed benchmark includes a broader set of questions and answers relating to medical concepts and also the differences between them, spanning various levels of difficulty. Furthermore, our evaluation encompasses not only general-purpose LLMs but also CLLMs specialized for the medical domain. Unlike their focus on a single hospital site, our medical coding evaluation covers a wider scope and also contains drugs medical concepts. Additionally, our evaluation also incorporates few-shot learning learning and not only zero-shot learning. 
% Moreover, Soroush et al. fed code descriptions to get the code which is a specific task that is currently being performed by manual coders. We wanted to evaluate the "clinical's understanding" of the LLMs and we did it by querying the other direction, from a clinical code to a description.
Moreover, Soroush et al. generated code from natural language descriptions (a task that is currently performed by human coders), while our study evaluates the "clinical understanding" of LLMs by investigate the inverse direction: selection the right description given a clinical code. In contrast to their approach, which assessed the capability of models to potentially replace human coders, our goal is to provide a more general estimation of the clinical reasoning abilities of LLMs.

% Our benchmark
In this study, we introduce MedConceptsQA, a benchmark for evaluating understanding and reasoning abilities in the medical domain. MedConceptsQA comprises over 800,000 questions and answers covering medical concepts, including ICD10 and ICD9 diagnoses codes, ICD9-PROC and ICD10-PROC procedures codes, and ATC drug codes. The benchmark includes questions categorized into three difficulty levels: easy, medium, and hard.

% Our Results for the benchmark
In addition, we evaluate different LLMs on the MedConceptsQA benchmark using zero-shot and few-shot learning. Surprisingly, we find that the current state-of-the-art CLLMs, as well as general LLMs except two, achieve performance levels comparable to Random guessing. Even medical fine-tuned models were not better than Random guessing. However, GPT-4 \cite{achiam2023gpt} outperforms all medical LLMs, despite its primary focus being a general LLM rather than specifically tailored to the medical domain like the others.

% Our contribution. 
\textbf{Contributions}: In summary, we make three main contributions: 

\begin{enumerate}
    \item We propose a challenging open-source benchmark for evaluating LLMs on their understanding and reasoning of medical concepts. Additionally, our evaluation code is also open-source and available for any use.
    \item Due to our large number of examples in the benchmark (more than 800,000), our benchmark can be utilized for training LLMs to comprehend medical concepts and the distinctions between them, making it suitable for techniques such as instruction-tuning \cite{zhang2023instruction}. 
    \item We demonstrate that GPT-3.5 and GPT-4 outperform the current state-of-the-art CLLMs in understanding and reasoning about medical concepts, despite CLLMs being specifically designed for the medical domain.
\end{enumerate}

\section{MedConceptsQA Benchmark}
The MedConceptsQA benchmark consists of multiple question-answering medical concepts datasets, categorized into three vocabularies: Diagnoses, Procedures, and Drugs. Within the Diagnoses category, it includes questions related to ICD9-CM and ICD10-CM medical codes. For Procedures, it covers ICD9-PROC and ICD10-PROC codes, and ATC for Drugs. Each vocabulary is further divided into three levels of difficulty: easy, medium, and hard. Table \ref{tab:benchmark_distribution} illustrates the distribution of questions within the benchmark.

\begin{table}[ht]
\centering
\begin{tabular}{cllr} % Adjust the alignment of the vocab column to 'c' for centering
\toprule
{} & \multicolumn{1}{c}{Vocab} &   Level &   Questions num \\ 
\midrule
\multirow{3}{*} & \multirow{3}{*}{ATC} &    easy &    6440 \\
                    &                                     &  medium &    6440 \\
                    &                                     &    hard &    5938 \\
\midrule
\multirow{3}{*} & \multirow{3}{*}{ICD10-CM} &    easy &   94580 \\
                    &                                                &  medium &   81757 \\
                    &                                                &    hard &   88013 \\
\midrule
\multirow{3}{*} & \multirow{3}{*}{ICD10-PROC} &    easy &  190987 \\
                    &                                                    &  medium &  190987 \\
                    &                                                    &    hard &   88582 \\
\midrule
\multirow{3}{*} & \multirow{3}{*}{ICD9-CM} &    easy &   17736 \\
                    &                                                &  medium &   17736 \\
                    &                                                &    hard &   16858 \\
\midrule
\multirow{3}{*} & \multirow{3}{*}{ICD9-PROC} &    easy &    4670 \\
                    &                                                    &  medium &    4670 \\
                    &                                                    &    hard &    4438 \\ 
\bottomrule
\end{tabular}
\caption{The number of generated questions across different vocabularies and difficulty levels in the benchmark.}
\label{tab:benchmark_distribution}
\end{table}

Each question is about a single medical code and contains four optional answers. The options are four descriptions of medical codes, where only one is the true description of the given medical codes, and the others are randomly chosen according to the category of difficulty of the specific level. The order of the options and the placement of correct answers are chosen randomly. 

\subsection{Difficulty Levels Creation}
\begin{figure}
    \centering
    \includegraphics[width=1\linewidth]{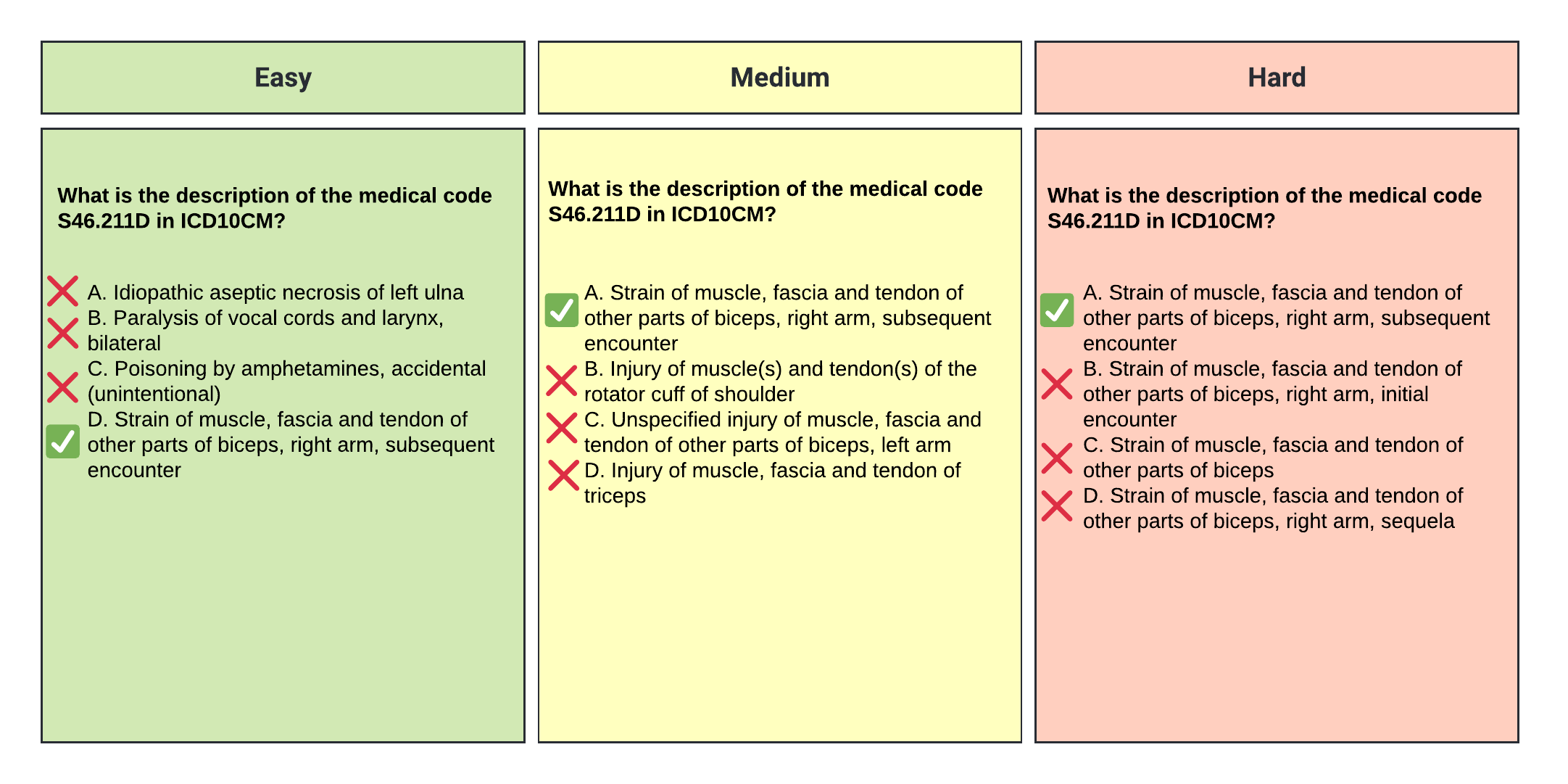}
    \caption{Questions across three difficulty levels for the S46.211D medical code within the ICD10-CM vocabulary.}
    \label{fig:questions_for_levels}
\end{figure}

We represent the medical code vocabulary hierarchy as an undirected graph using PyHealth \cite{pyhealth2023yang}. Each vocabulary is divided into three difficulty levels: easy, medium, and hard. The distinction lies in the options provided for the correct answer. For the easy level, the options are randomly chosen from all the medical concept codes within the vocabulary. For the medium level, we select alternative medical codes within a distance of three, four, or five edges from each node in the vocabulary. For the hard level, the required distance is reduced to two edges. This means that for the hard level, the candidates are closely related, sharing a common parent or a node and its grand-parent or a node and its grand-child. For example, one of the candidates for the medical code S46.211D in ICD10-CM is S46.212A, because these codes share a common parent. Figure \ref{fig:questions_for_levels} presents examples of questions for the same ICD10-CM code across each difficulty level. 
Algorithm \ref{alg:generate_questions_algo} presents the pseudo-code we used to generate the questions based on the required distance for each level. We used this algorithm to create the medium and hard questions. Easy questions were generated by a uniform sampling of the codes across the entire codes' graph.

\begin{algorithm}
\caption{Generating Questions by Distance}
\begin{algorithmic}[1]
\Function{createQuestionsByDistance}{$\text{vocab\_graph}, \text{required\_edges\_distance}, \text{num\_of\_options}$}
\State \textbf{Input:} Vocabulary graph $\text{vocab\_graph}$, Required edges distance: $\text{required\_edges\_distance}$, Number of options: $\text{num\_of\_options}$
\State \textbf{Output:} List of generated questions
\State $\text{nodes} \gets \text{vocab\_graph.nodes}$
\State $\text{questions} \gets \varnothing$
\For{$\text{node}$ \textbf{in} $\text{nodes}$}
\State $\text{candidate\_nodes} \gets \varnothing$
\For{$\text{candidate\_node}$ \textbf{in} $\text{nodes} \setminus \{\text{node}\}$}
\State $\text{path\_length} \gets \text{get\_shortest\_path\_length}(\text{node}, \text{candidate\_node})$
\If{$\text{path\_length}$ \textbf{in} $\text{required\_edges\_distance}$}
\State $\text{candidate\_nodes} \gets \text{candidate\_nodes} \cup \{\text{candidate\_node}\}$
\EndIf
\EndFor
\State $\text{candidates\_number} \gets \text{length of candidate\_nodes}$
\If{$\text{candidates\_number} \geq \text{num\_of\_options}$}
\State $\text{options} \gets \text{choose\_random(candidate\_nodes, num\_of\_options)}$
\State $\text{question} \gets \text{generate\_question(node, options)}$
\State $\text{questions} \gets \text{questions} \cup \{\text{question}\}$
\EndIf
\EndFor
\State \textbf{return} $\text{questions}$
\EndFunction
\end{algorithmic}
\end{algorithm}
\label{alg:generate_questions_algo}

\section{Experiments}
% Experiments setup. 
In this section, we evaluate CLLMs on the MedConceptsQA benchmark. The CLLMs we use for evaluations are: BioMistral (BioMistral-7B-DARE) \cite{labrak2024biomistral}, BioMedGPT (BioMedGPT-LM-7B) \cite{luo2023biomedgpt}, Gatortron (gatortron-large) \cite{yang2022large}, Llama3-OpenBioLLM (70B) \cite{OpenBioLLMs}, BioBert \cite{lee2020biobert}, Meerket (7B) \cite{kim2024small}, Meditron (70B) \cite{chen2023meditron}, Clinical-Longformer \cite{li2023comparative}. Additionally, we include GPT-3.5 and GPT-4 \cite{achiam2023gpt} for our evaluation, although the main focus of these models is to be general-purpose LLMs and not specifically focused on a clinical domain.

\begin{figure}[ht]
    \centering
    \begin{subfigure}[b]{0.45\linewidth}
        \centering
        \includegraphics[width=\linewidth]{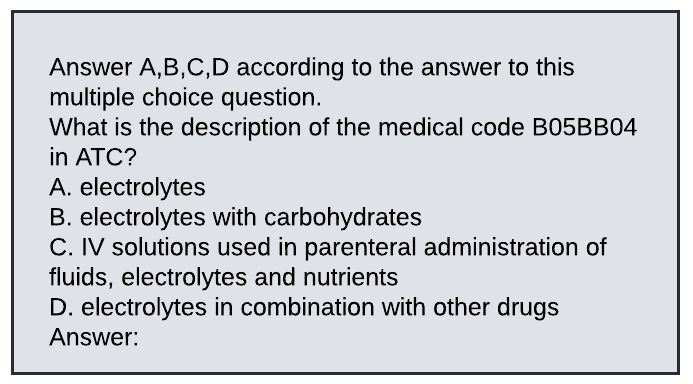}
        \caption{Zero-shot learning prompt example, for ATC vocabulary.}
        \label{fig:zero-shot-example}
    \end{subfigure}
    \hfill
    \begin{subfigure}[b]{0.45\linewidth}
        \centering
        \includegraphics[width=\linewidth]{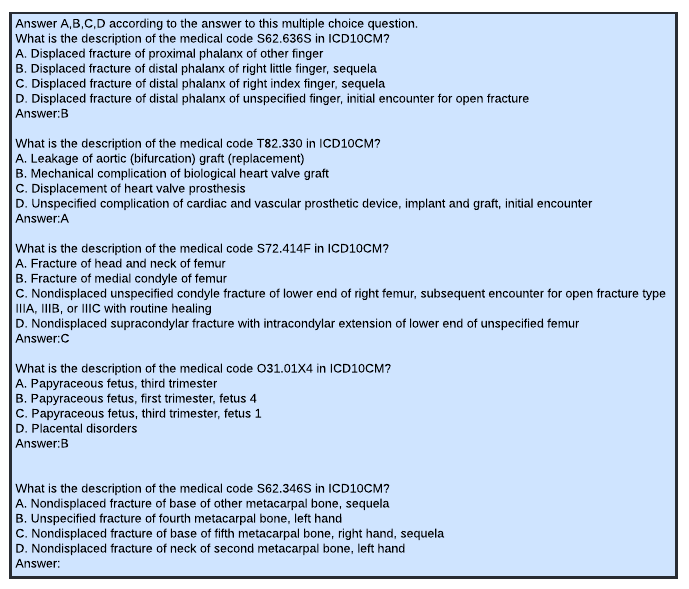}
        \caption{Few-shot learning prompt example, for ICD10-CM vocabulary. We used 4 shots, each one contains the question and the answer.}
        \label{fig:few-shot-example}
    \end{subfigure}
    \caption{Examples of zero-shot and few-shot learning prompts.}
    \label{fig:zero-and-few-shot-examples}
\end{figure}

Our evaluation is based on zero-shot learning and few-shot learning for all the models. For zero-shot learning we contain the question and general instruction description. For few-shot learning we include 4 shots in the prompt and then the question we want to get the answer for. Figure \ref{fig:zero-shot-example} shows an example for zero-shot learning prompt for our drugs dataset (ATC) in the benchmark, and figure \ref{fig:few-shot-example} shows an example for few-shot learning example from our benchmark, for ICD10-CM vocabulary.

\subsection{Experiments Setup}
For each dataset in the benchmark, we conduct zero-shot and few-shot evaluations at each difficulty level (easy, medium, and hard). We repeat each evaluation for each model three times and calculate a 95\% confidence interval. Due to the large amount of resources required by some of the models, especially GPT4, we limit each type of test (vocabulary and difficulty and shots) to 250 randomly sampled Q\&As. Few-shot learning is performed using 4-shot learning. All models except GPT-3.5 (gpt-3.5-turbo) and GPT-4 (gpt-4-0125-preview) were evaluated using HuggingFace \cite{wolf2019huggingface} on RTX-6000 GPU, while GPT-3.5 and GPT-4 were inferred using the OpenAI API because these models are not provided as open-source. We use accuracy as the evaluation metric because the datasets in the benchmark are balanced, as we selected the placement of the correct answer randomly during the creation of the benchmark.

\subsection{Aggregated Analysis}

\begin{figure}[ht]
    \centering
    \begin{subfigure}[b]{0.5\linewidth}
        \centering
        \includegraphics[width=\linewidth]{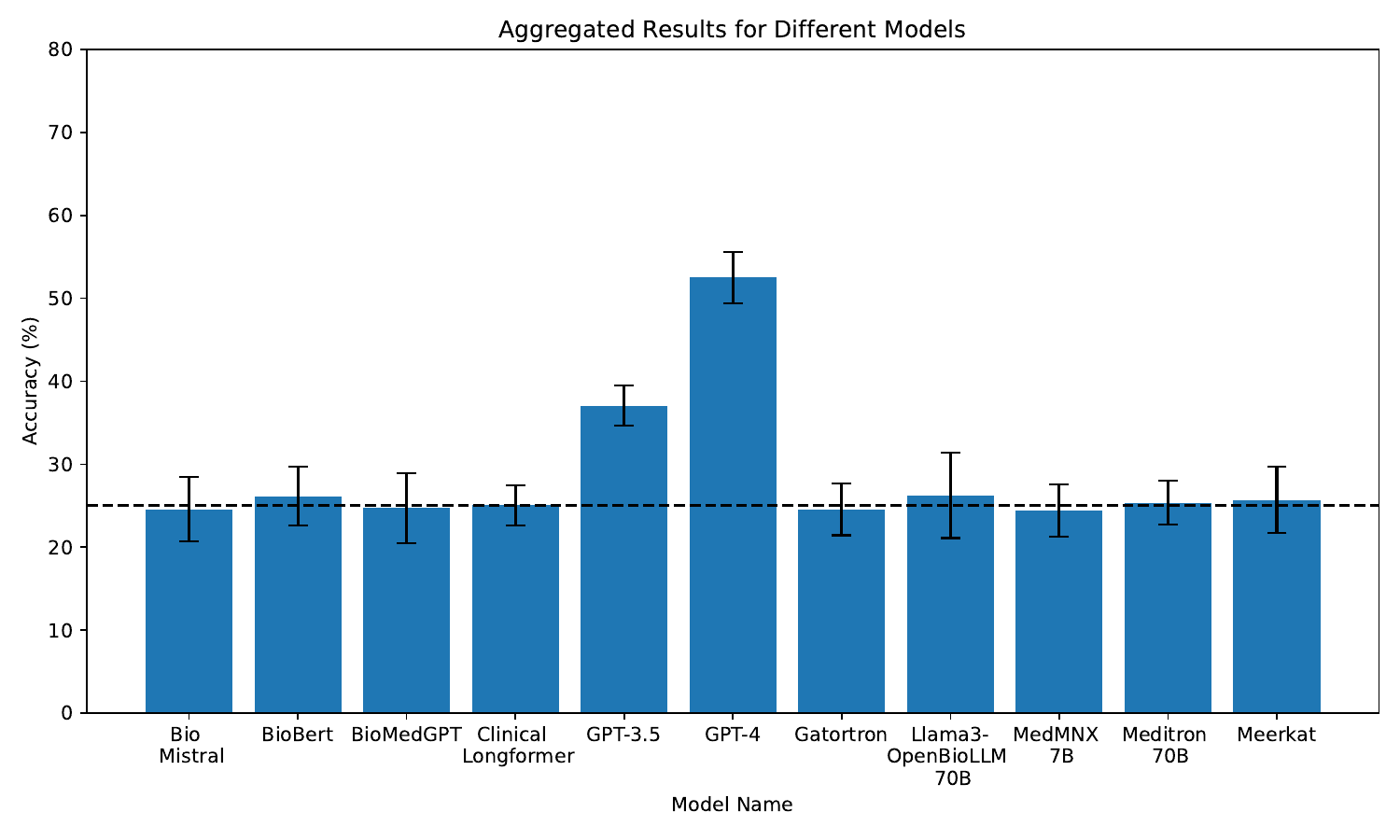}
        \caption{Zero-shot aggregated results for each model.}
        \label{fig:zero-shot-aggregated-results}
    \end{subfigure}%
    \begin{subfigure}[b]{0.5\linewidth}
        \centering
        \includegraphics[width=\linewidth]{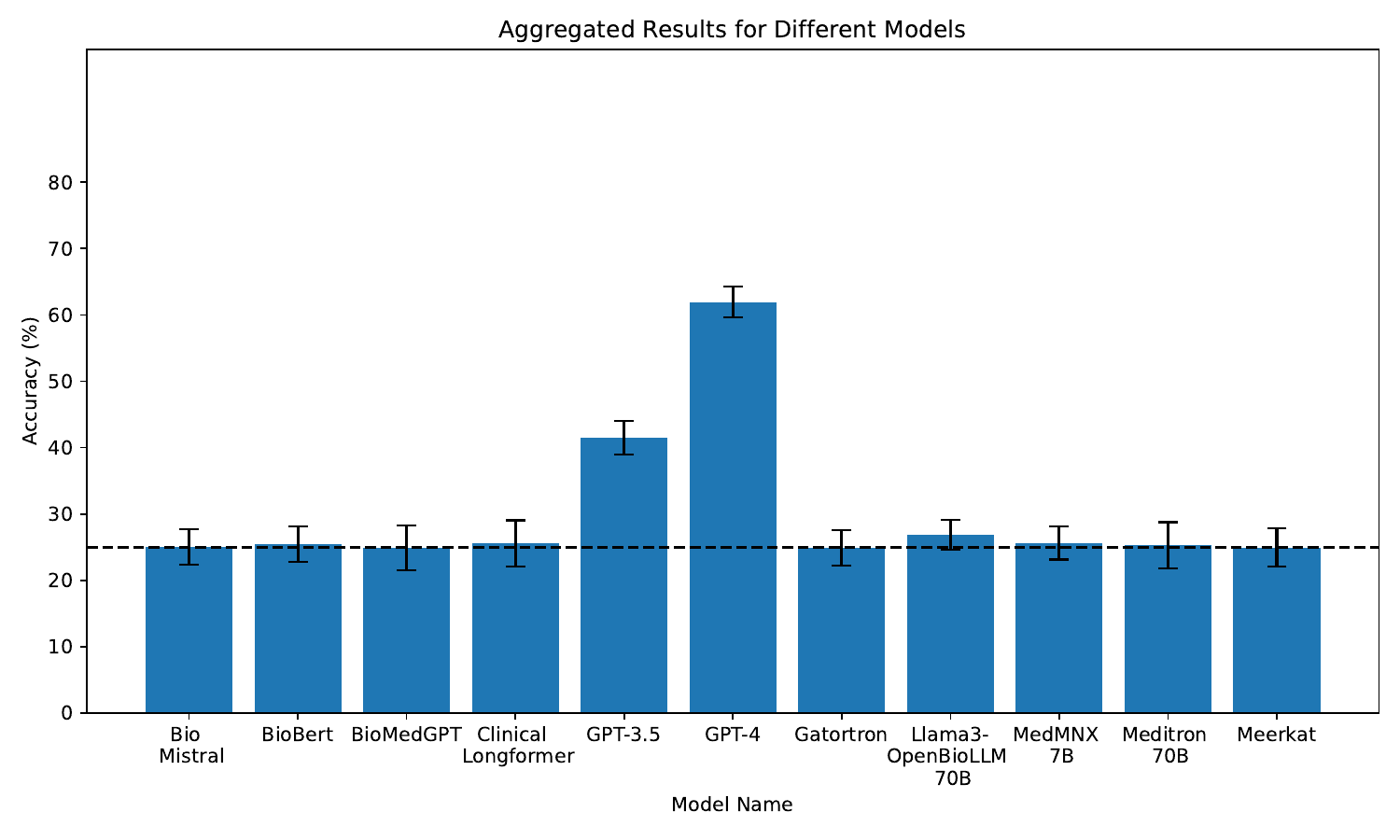}
        \caption{Few-shot aggregated results for each model.}
        \label{fig:few-shot-aggregated-results}
    \end{subfigure}
    \caption{Aggregated results for zero-shot and few-shot evaluations. The vertical line represents the accuracy of random guessing for comparison.}
    \label{fig:combined-results}
\end{figure}

We perform zero-shot and few-shot evaluations for each model across all vocabularies and difficulty levels (easy, medium, hard). In this section, we present aggregated results for each model. The aggregation includes averaging both the accuracy and the 95\% confidence interval. Figure \ref{fig:combined-results} presents the results of zero-shot and few-shot learning evaluation. The full results across all the models, vocabularies, and difficulty levels can be found in the appendix (Table \ref{tab:zs_full_results} for zero-shot learning and Table \ref{tab:zs_full_results} for few-shot learning).

The CLLMs achieve performance close to Random guessing. For zero-shot learning evaluation, None of the CLLMs outperforms the Random guessing threshold when considering the error rate (confidence intervals). However, both GPT-3.5 and GPT-4 outperform the CLLMs, with GPT-4 achieving the highest performance at 52.49\% accuracy. This result represents a significant improvement over random guessing, with an absolute increase of 27.49\%. In the case of few-shot learning, the results are similar for the CLLMs. However, GPT-3.5 shows an improvement in accuracy of 4.418\% (absolute), and GPT-4 demonstrates absolute accuracy improvements of 9.422\% between zero-shot and few-shot evaluations.

\subsection{Results for Difficulty Level}

\begin{figure}[ht]
    \centering
    \begin{subfigure}[b]{0.5\linewidth}
        \centering
        \includegraphics[width=\linewidth]{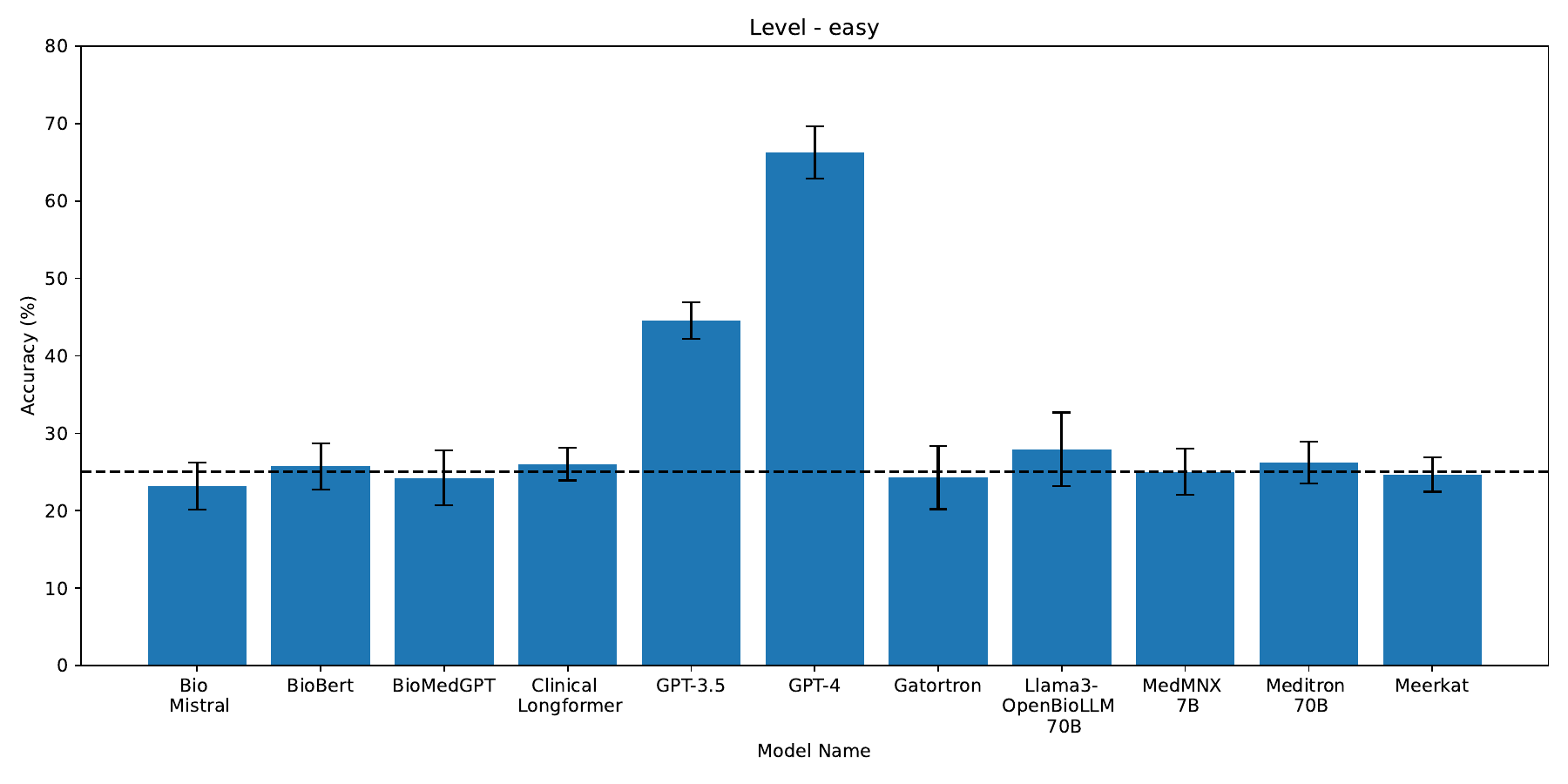}
        \caption{Easy level, zero-shot learning.}
        \label{fig:easy-zero-shot}
    \end{subfigure}%
    \begin{subfigure}[b]{0.5\linewidth}
        \centering
        \includegraphics[width=\linewidth]{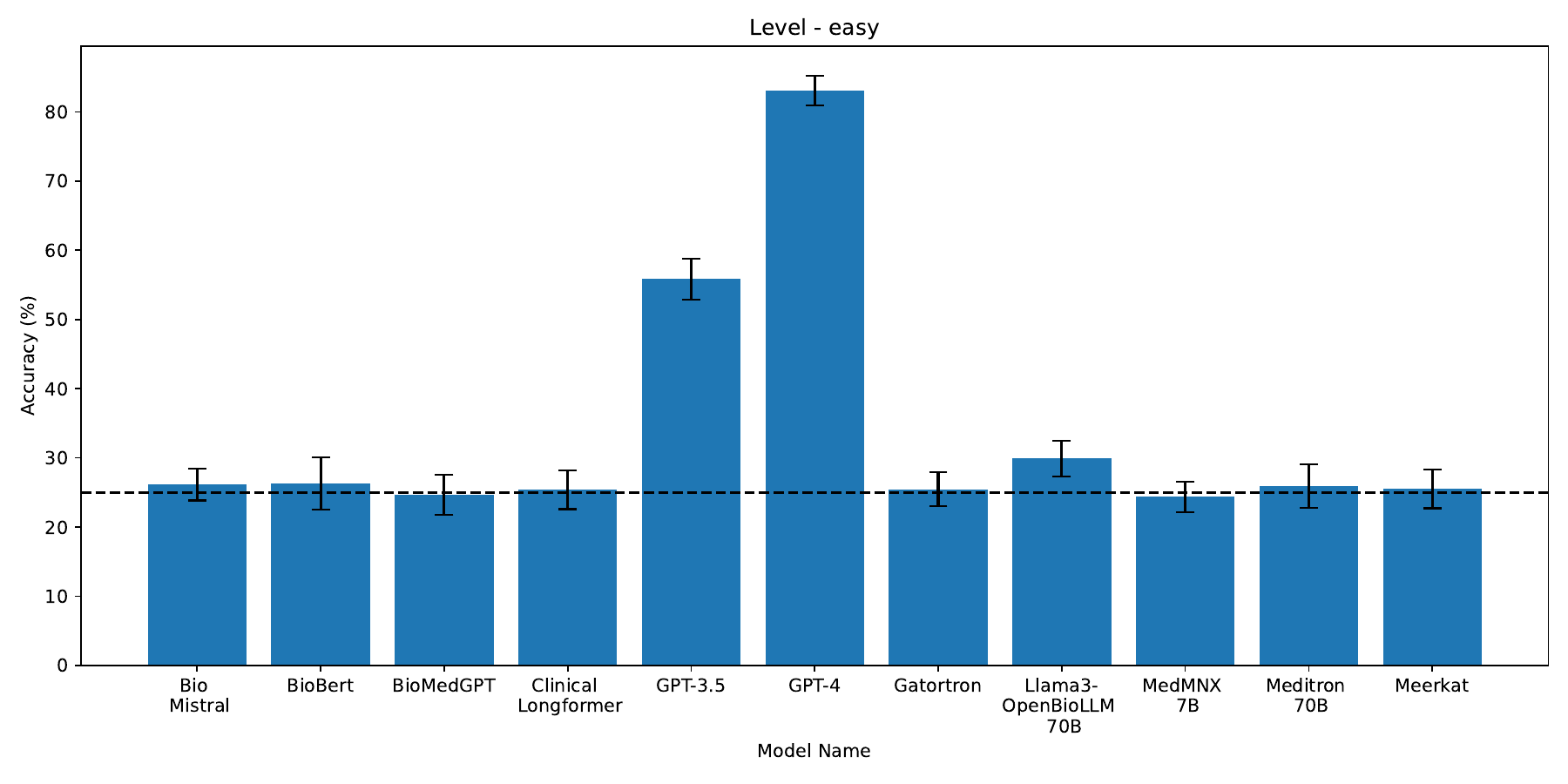}
        \caption{Easy level, few-shot learning.}
        \label{fig:easy-few-shot}
    \end{subfigure}

    \begin{subfigure}[b]{0.5\linewidth}
        \centering
        \includegraphics[width=\linewidth]{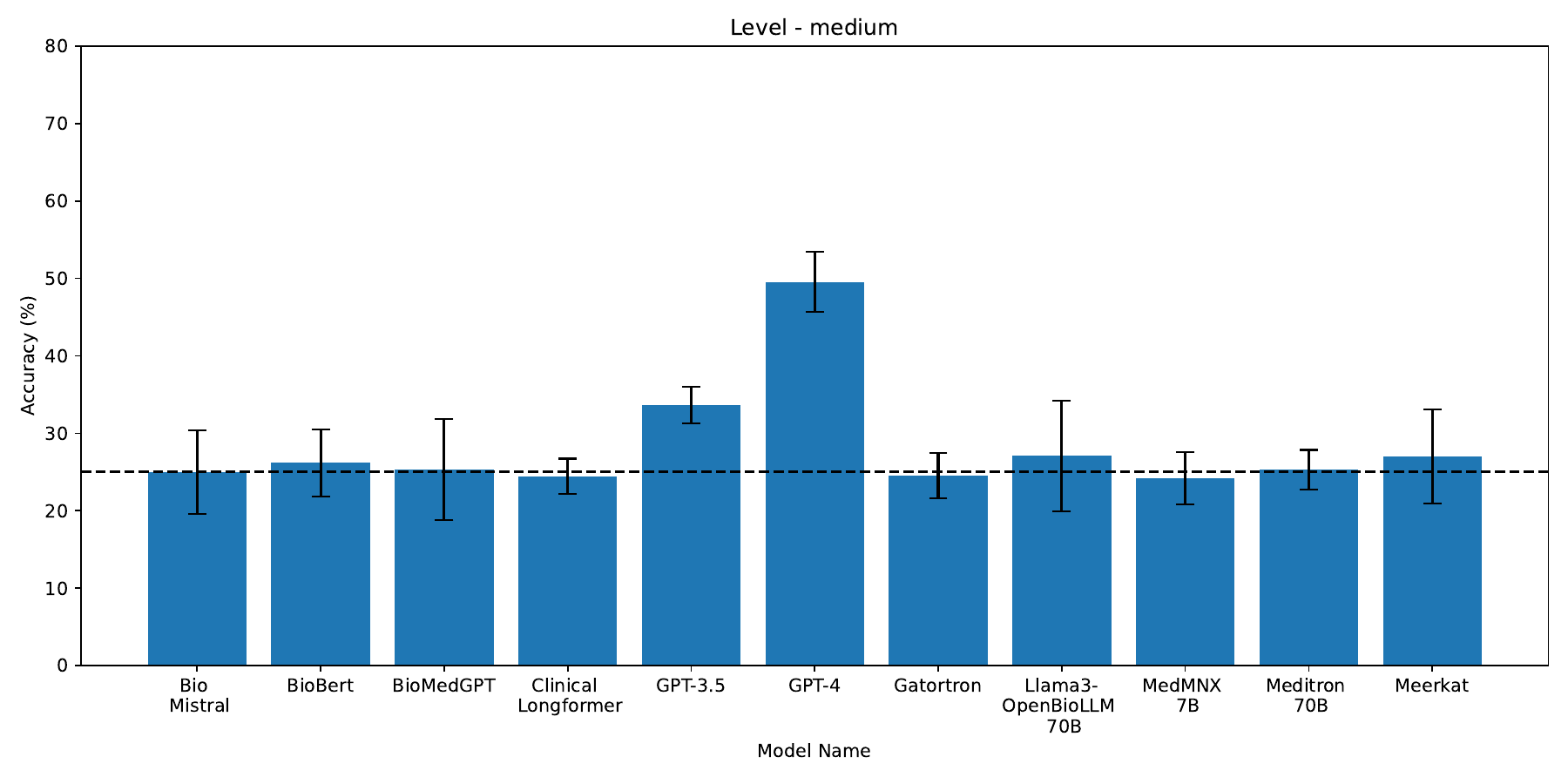}
        \caption{Medium level, zero-shot learning.}
        \label{fig:medium-zero-shot}
    \end{subfigure}%
    \begin{subfigure}[b]{0.5\linewidth}
        \centering
        \includegraphics[width=\linewidth]{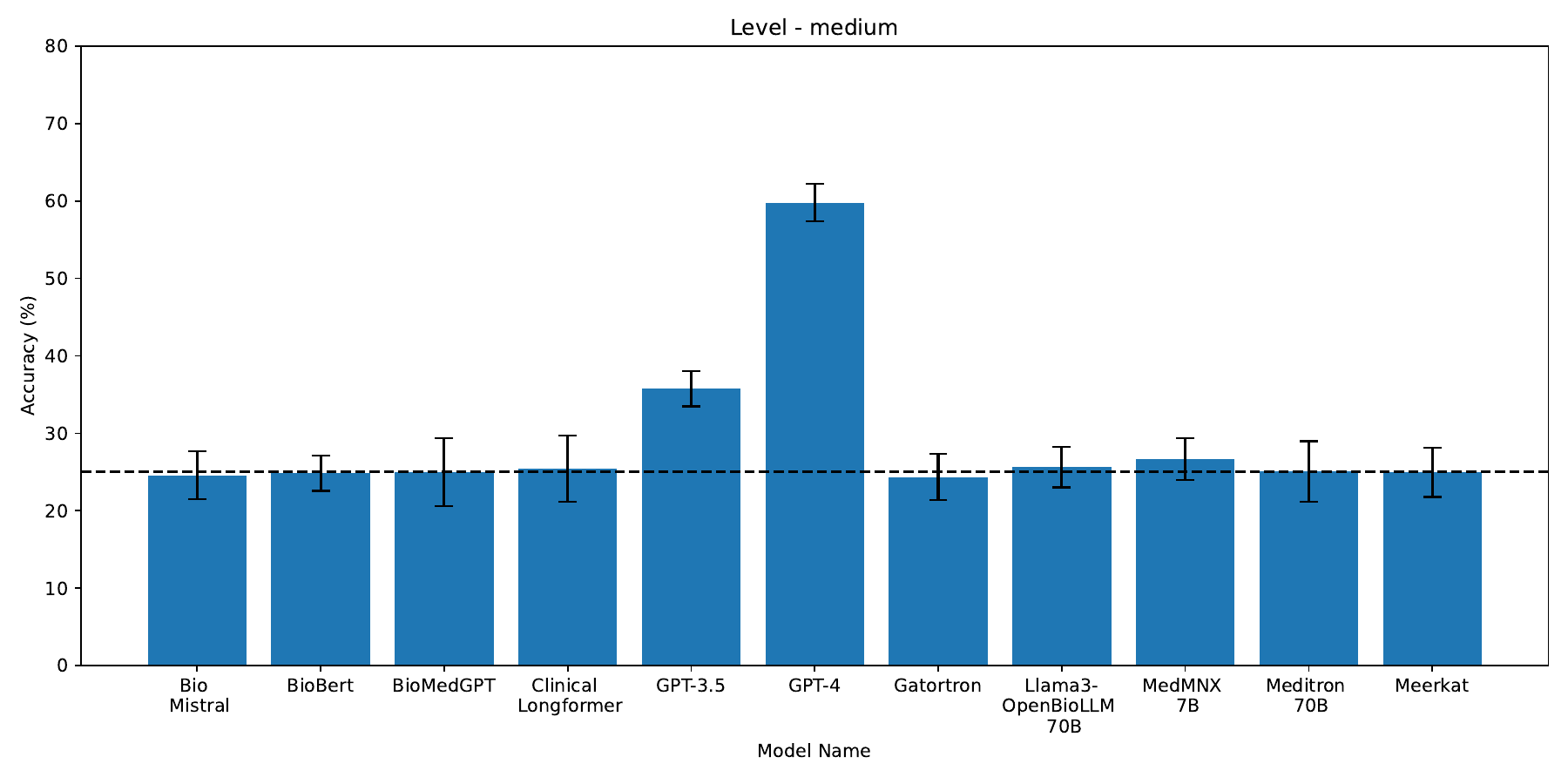}
        \caption{Medium level, few-shot learning.}
        \label{fig:medium-few-shot}
    \end{subfigure}

    \begin{subfigure}[b]{0.5\linewidth}
        \centering
        \includegraphics[width=\linewidth]{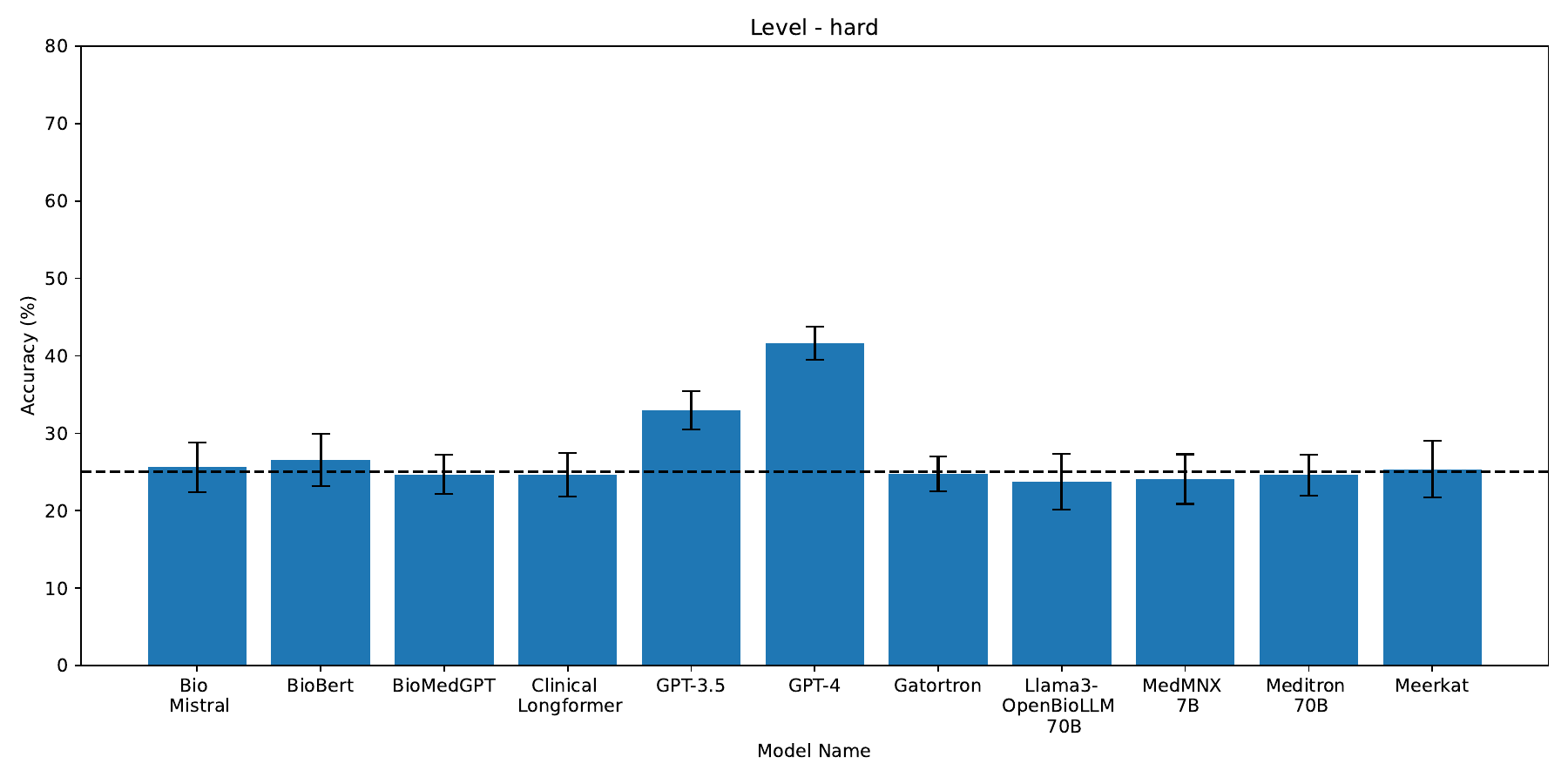}
        \caption{Hard level, zero-shot learning.}
        \label{fig:hard-zero-shot}
    \end{subfigure}%
    \begin{subfigure}[b]{0.5\linewidth}
        \centering
        \includegraphics[width=\linewidth]{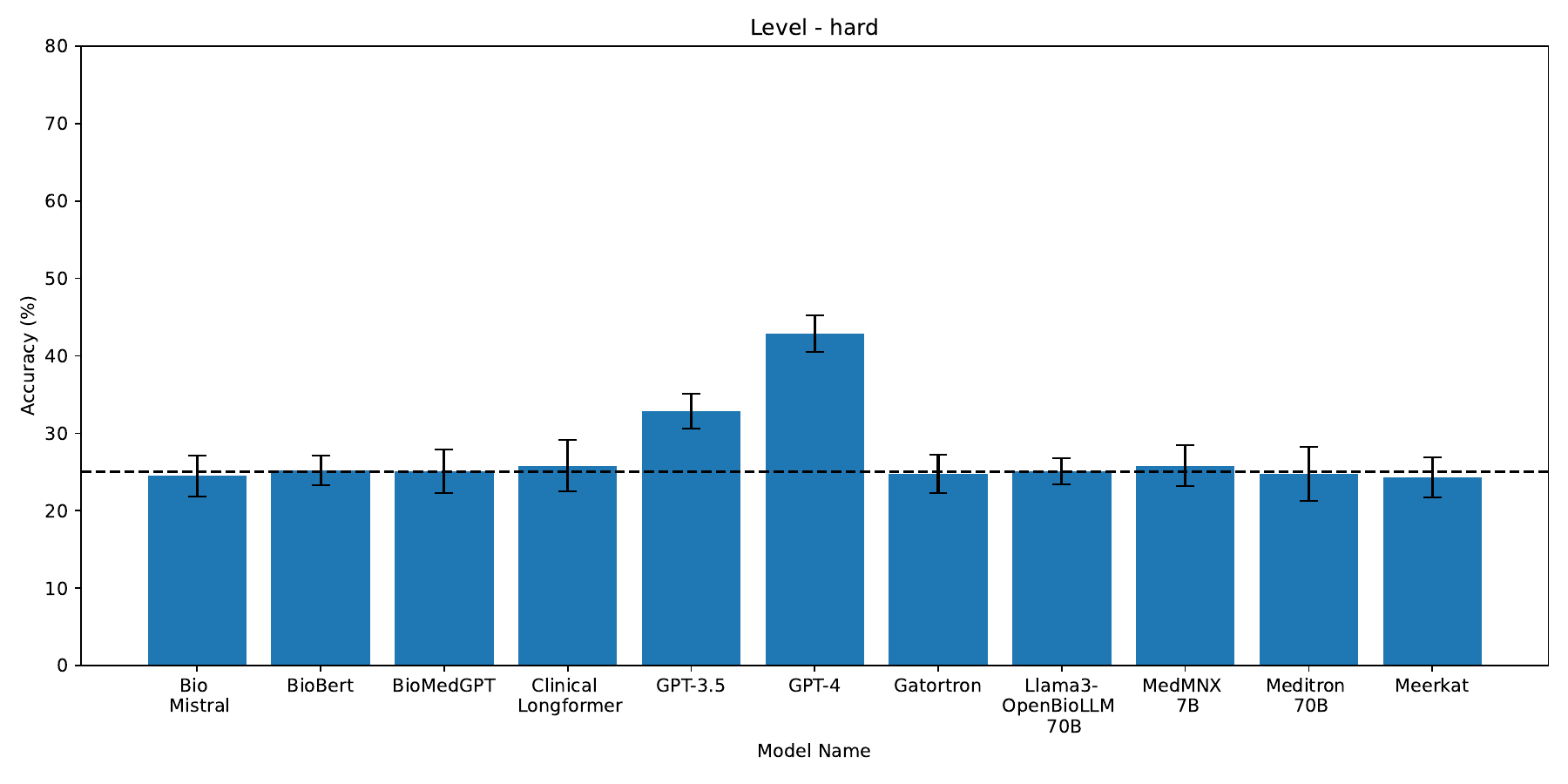}
        \caption{Hard level, few-shot learning.}
        \label{fig:hard-few-shot}
    \end{subfigure}
    \caption{Zero-shot and few-shot results for each of the levels (easy, medium, and hard) with 95\% confidence intervals over three runs. Results are aggregated over difficulty vocabularies.}
    \label{fig:levels_results}
\end{figure}

We create three difficulty levels (easy, medium, and hard) for each dataset in our benchmark. Figure \ref{fig:levels_results} shows the results of all models across the different levels for both zero-shot and few-shot learning settings. Nearly all the CLLMs achieve random guessing accuracy across levels (considering error bars), except for LLaMA-3B OpenBioLLM-70B, which performs better than random guessing on the easy level under few-shot learning. GPT-3.5 and GPT-4 demonstrate improvement from zero-shot to few-shot learning, indicating that the 4-shot examples enhance their performance. Moreover, the performance of GPT-3.5 and GPT-4 deteriorates as task difficulty increases, validating that our benchmark levels indeed represent a gradual increase in complexity. In Section \ref{sec:case_study_icd9}, we present a case study focusing on the ICD9-CM vocabulary and analyze results across difficulty levels.

\subsection{Results for Different Vocabularies}
\begin{figure}[ht]
    \centering
    \begin{subfigure}[b]{0.5\linewidth}
        \centering
        \includegraphics[width=\linewidth]{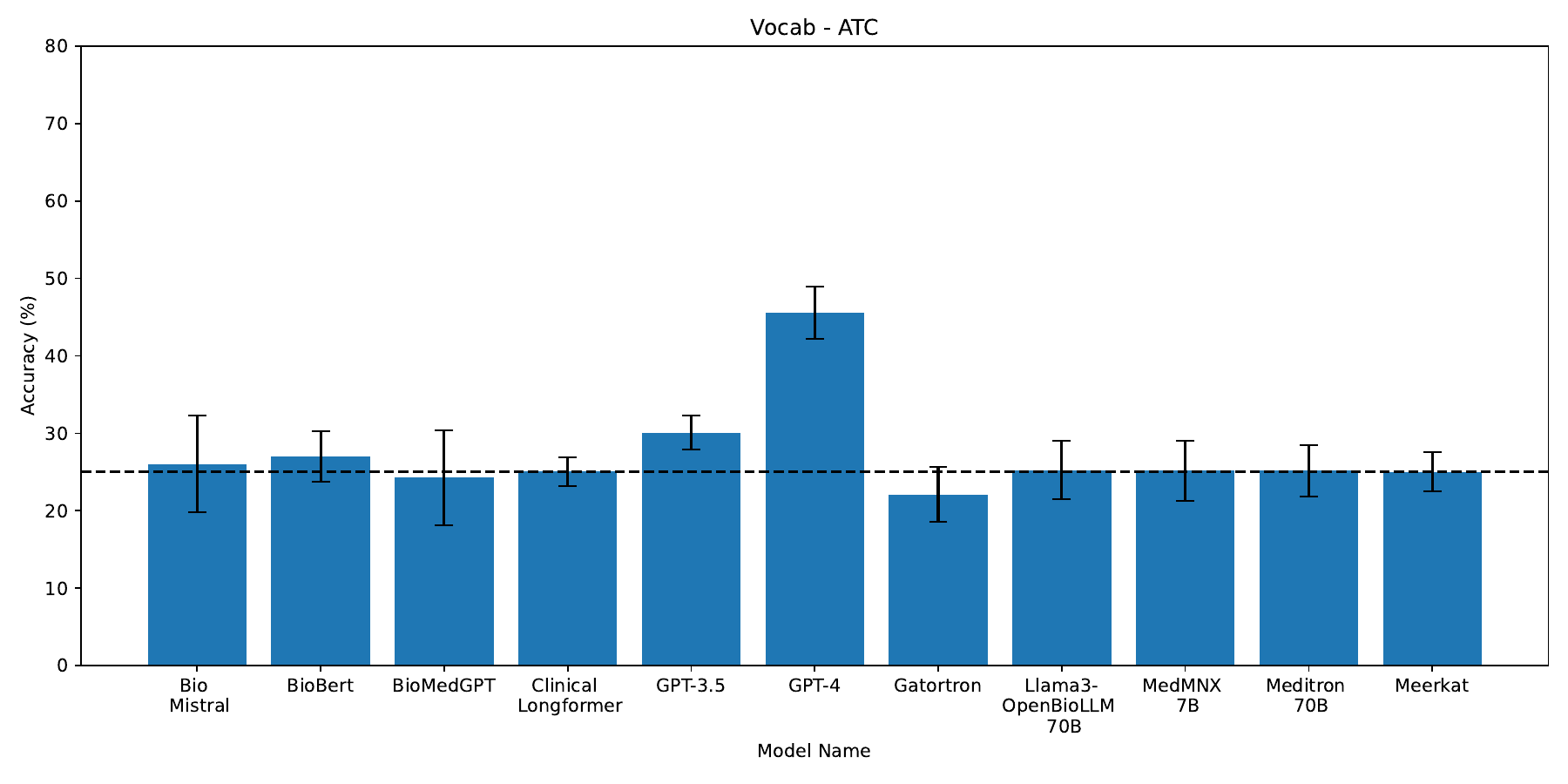}
        \caption{ATC vocabulary, zero-shot}
        \label{fig:atc-zero-shot}
    \end{subfigure}%
    \begin{subfigure}[b]{0.5\linewidth}
        \centering
        \includegraphics[width=\linewidth]{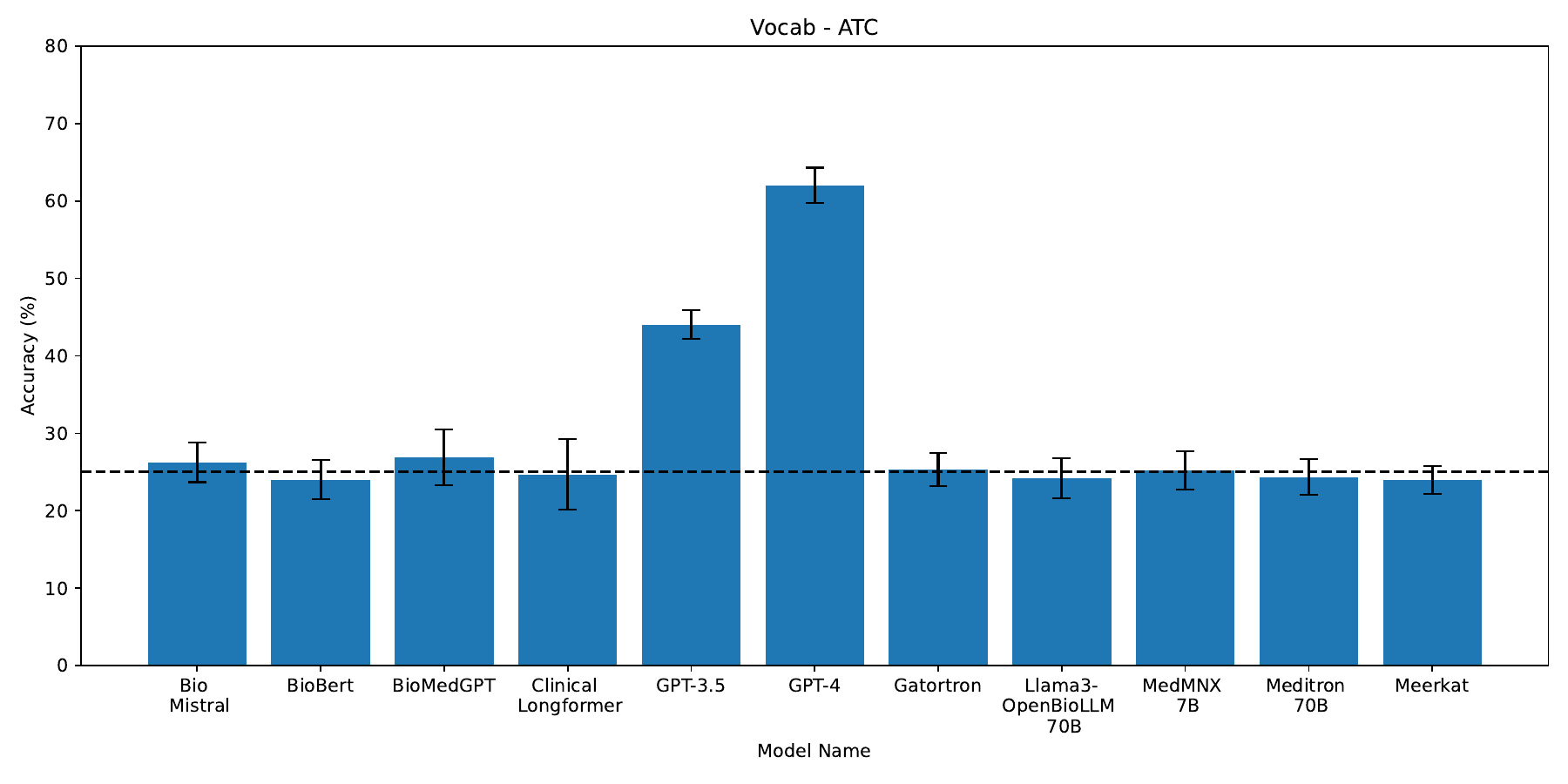}
        \caption{ATC vocabulary, few-shot}
        \label{fig:atc-few-shot}
    \end{subfigure}

    \begin{subfigure}[b]{0.5\linewidth}
        \centering
        \includegraphics[width=\linewidth]{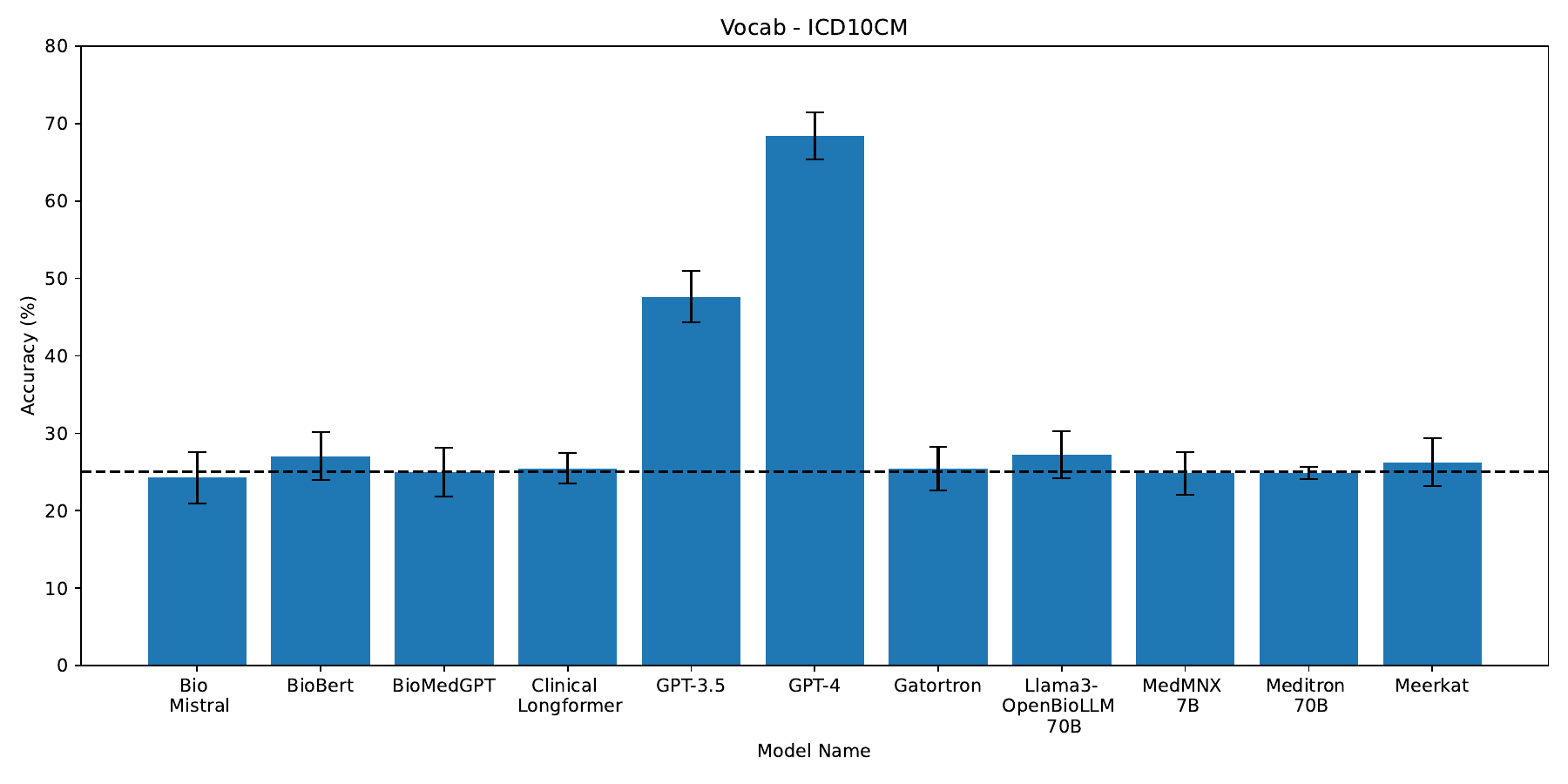}
        \caption{ICD10-CM vocabulary, zero-shot}
        \label{fig:ICD10-CM-zero-shot}
    \end{subfigure}%
    \begin{subfigure}[b]{0.5\linewidth}
        \centering
        \includegraphics[width=\linewidth]{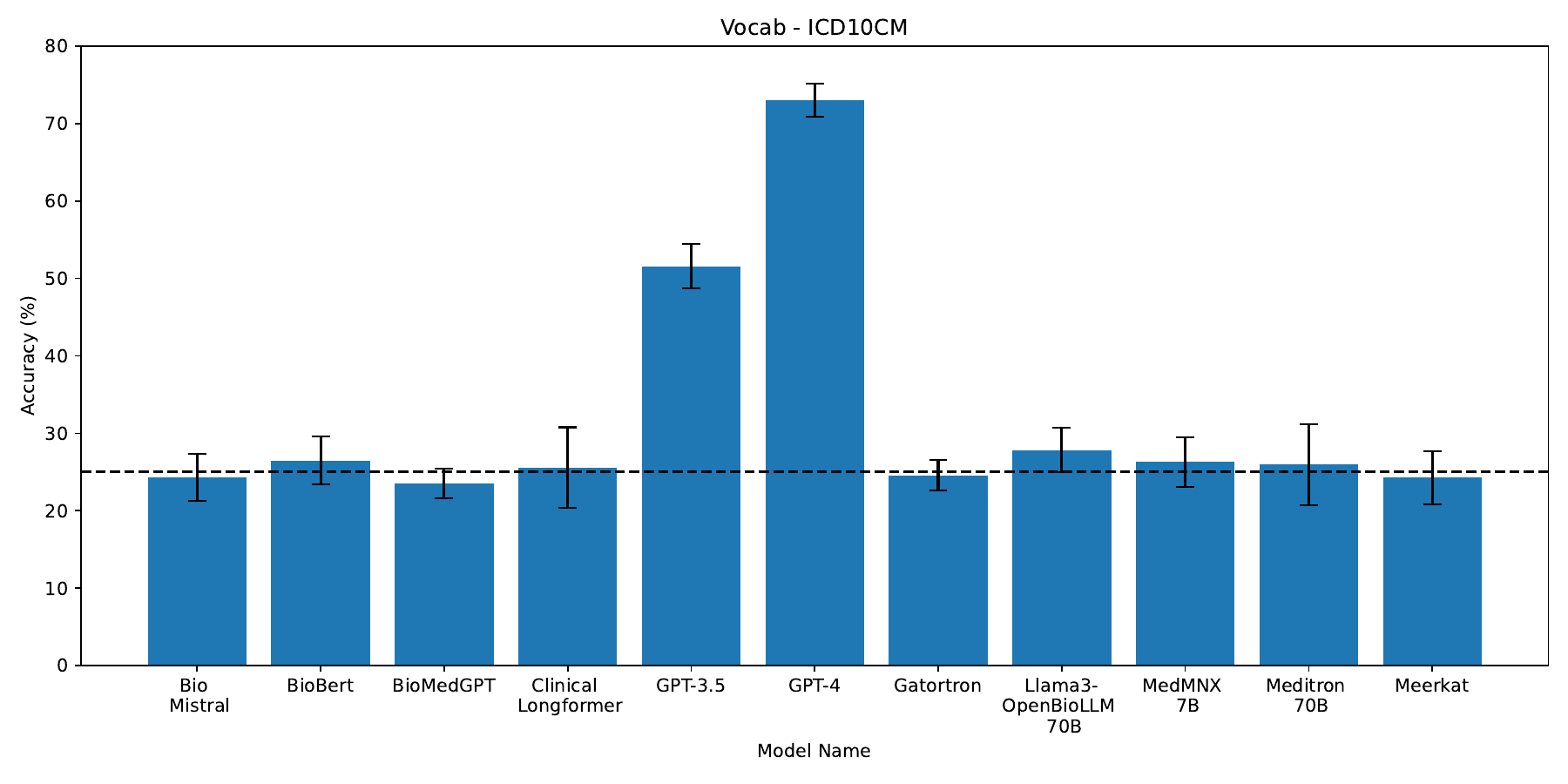}
        \caption{ICD10-CM vocabulary, few-shot}
        \label{fig:ICD10-CM-few-shot}
    \end{subfigure}

    \begin{subfigure}[b]{0.5\linewidth}
        \centering
        \includegraphics[width=\linewidth]{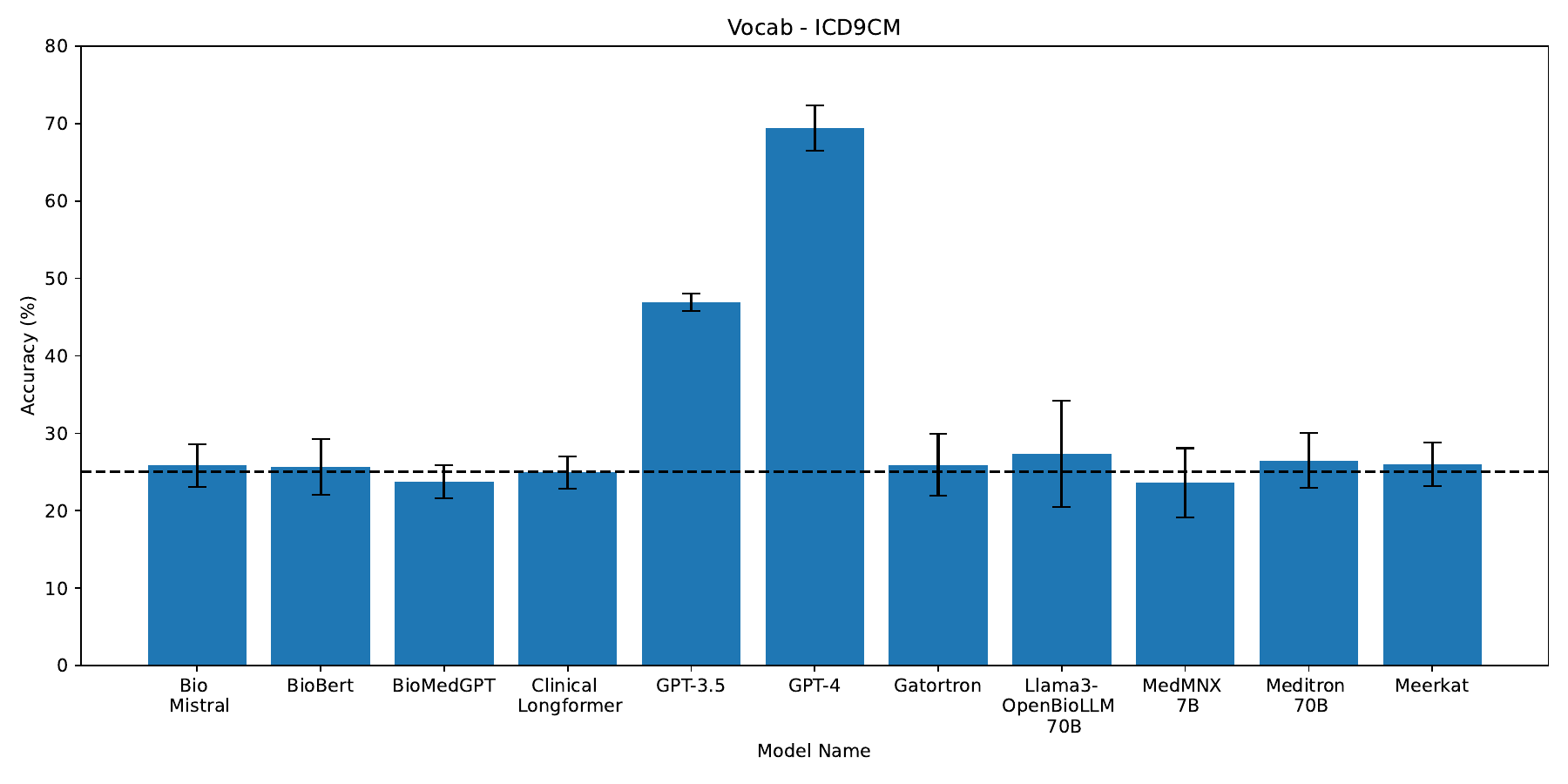}
        \caption{ICD9-CM vocabulary, zero-shot}
        \label{fig:ICD9-CM-zero-shot}
    \end{subfigure}%
    \begin{subfigure}[b]{0.5\linewidth}
        \centering
        \includegraphics[width=\linewidth]{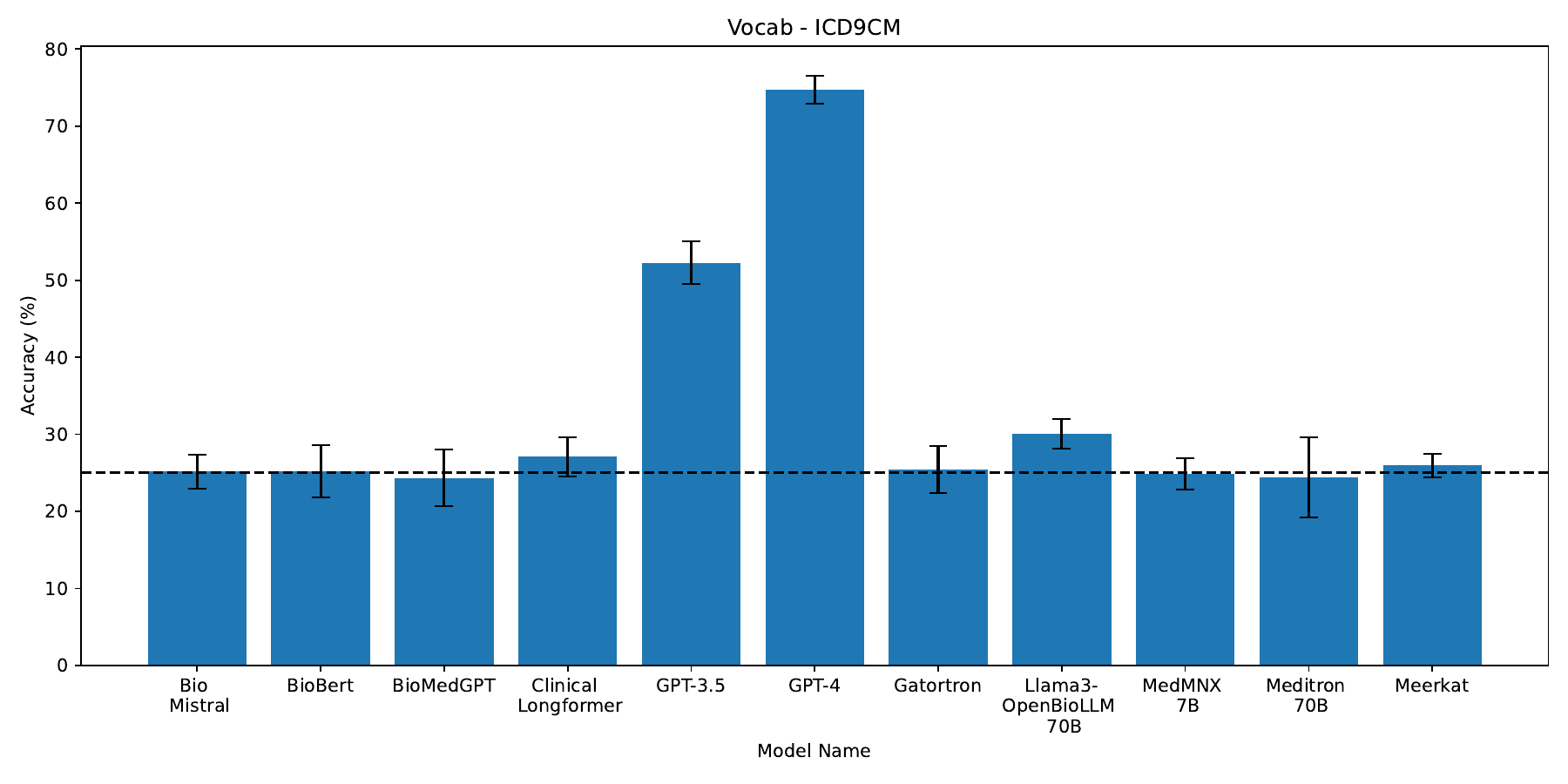}
        \caption{ICD9-CM vocabulary, few-shot}
        \label{fig:ICD9-CM-few-shot}
    \end{subfigure}

    \begin{subfigure}[b]{0.5\linewidth}
        \centering
        \includegraphics[width=\linewidth]{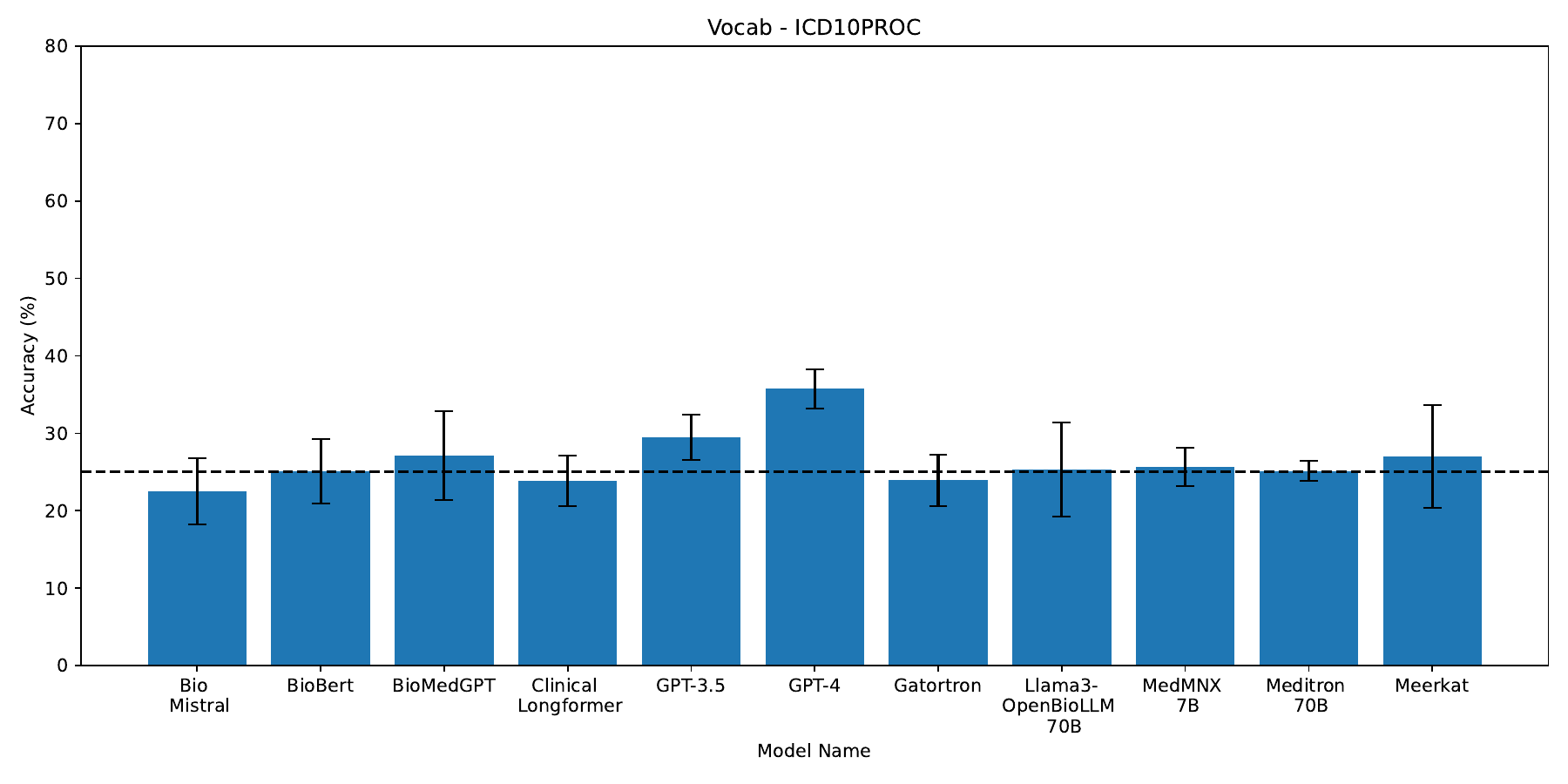}
        \caption{ICD10-PROC vocabulary, zero-shot}
        \label{fig:ICD10-PROC-zero-shot}
    \end{subfigure}%
    \begin{subfigure}[b]{0.5\linewidth}
        \centering
        \includegraphics[width=\linewidth]{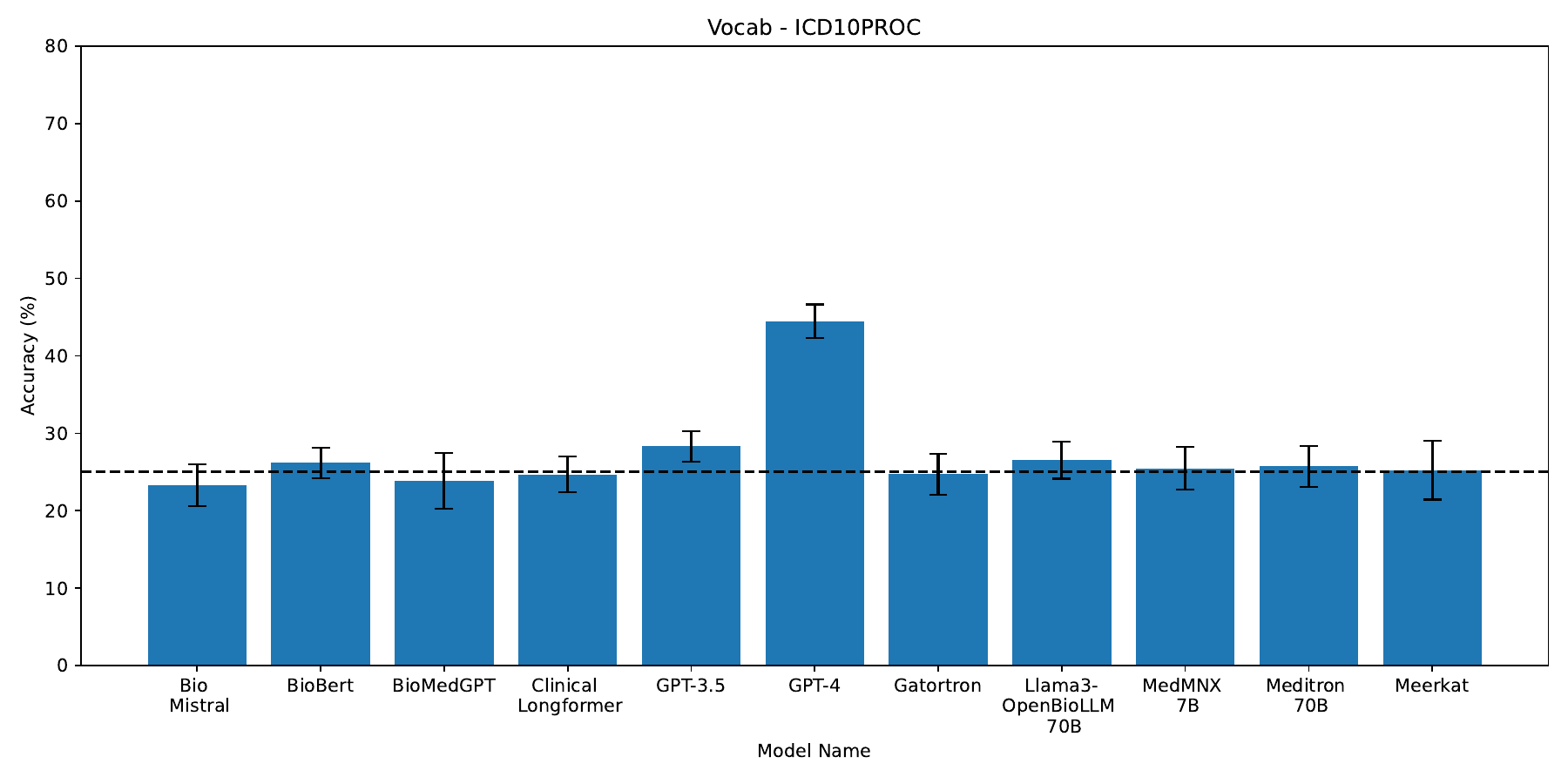}
        \caption{ICD10-PROC vocabulary, few-shot}
        \label{fig:ICD10-PROC-few-shot}
    \end{subfigure}

    \begin{subfigure}[b]{0.5\linewidth}
        \centering
        \includegraphics[width=\linewidth]{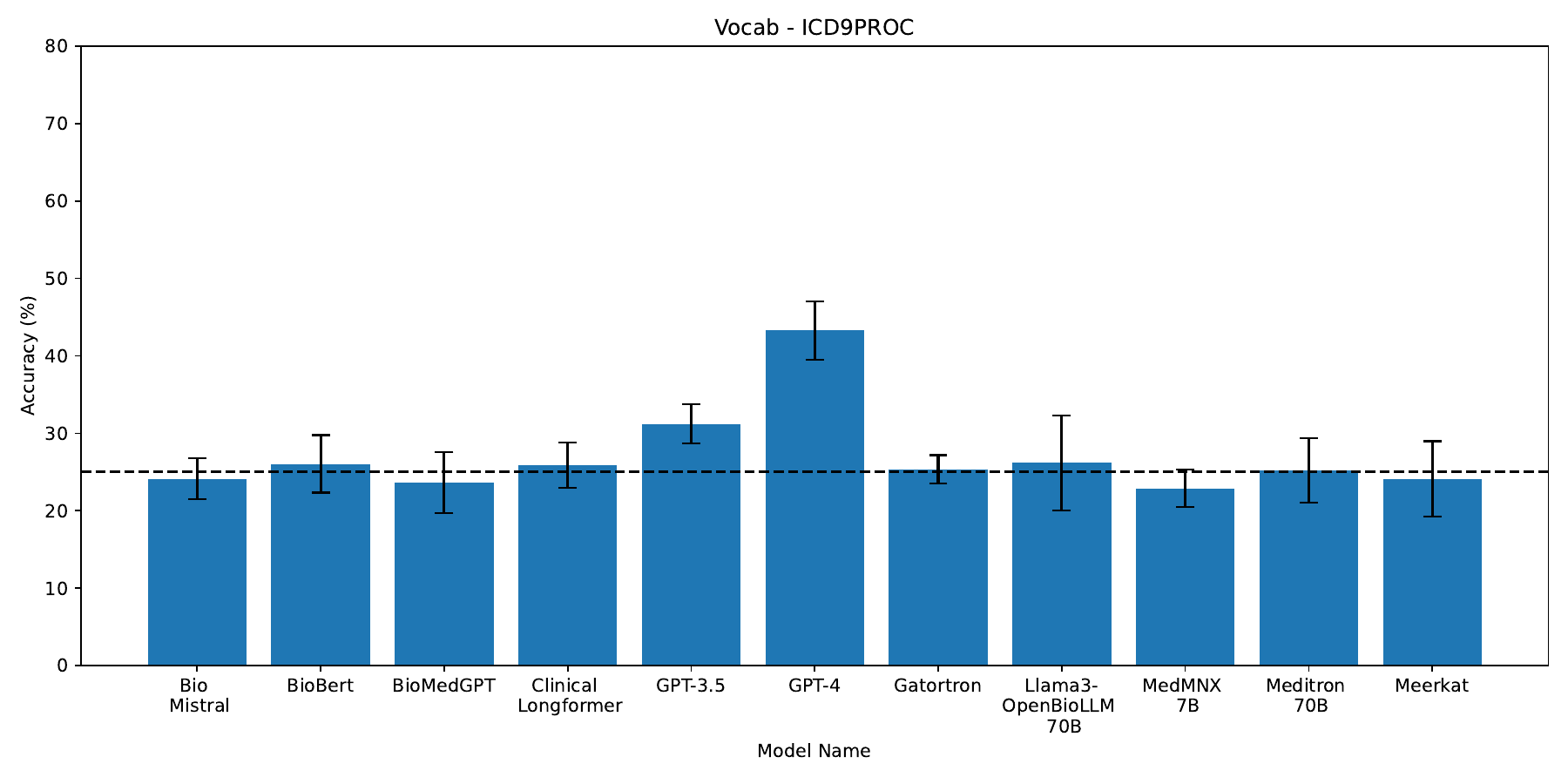}
        \caption{ICD9-PROC vocabulary, zero-shot}
        \label{fig:ICD9-PROC-zero-shot}
    \end{subfigure}%
    \begin{subfigure}[b]{0.5\linewidth}
        \centering
        \includegraphics[width=\linewidth]{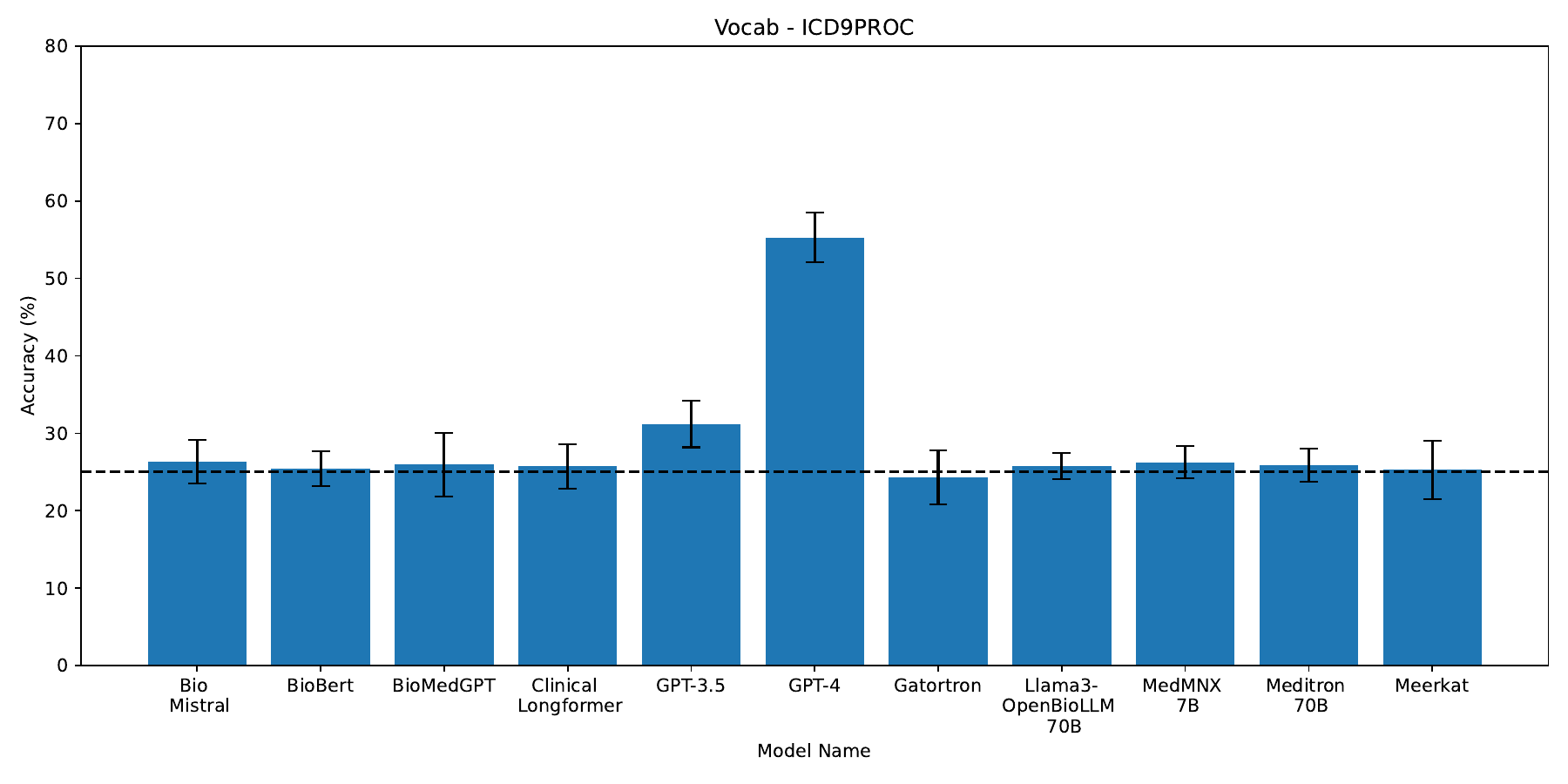}
        \caption{ICD9-PROC vocabulary, few-shot}
        \label{fig:ICD9-PROC-few-shot}
    \end{subfigure}

    \caption{Zero-shot and few-shot results for each of the vocabularies with 95\% confidence intervals over three runs. Results are aggregated over difficulty levels.}
    \label{fig:vocab_results}
\end{figure}

The accuracy results for each vocabulary (ATC, ICD9-CM, ICD10-CM, ICD9-PROC, ICD10-PROC) are presented in Figure \ref{fig:vocab_results}. The aggregation for the accuracy and 95\% CI is for all the levels for each vocabulary. Also here, GPT-4 consistently outperform all other models across all vocabularies, with GPT-3.5 in second place. The CLLMs achieve performance close to random guessing. For ATC, GPT-4 get an accuracy of 45.6\% with zero-shot learning but 62.045\% with few-shot learning. In the case of ICD10-CM, GPT-4 demonstrates the highest accuracy at 73.022\% with few-shot learning, while for ICD9-CM, it achieves 74.711\%. For procedures medical codes (ICD9-PROC and ICD10-PROC), the performance is lower than the other vocabularies. The best model is also GPT-4 for both few-shot evaluations, achieving scores of 44.489\% for ICD10-PROC and 55.289\% for ICD9-PROC. Overall, these experiments show that almost all CLLMs achieve performance close to random guessing, not only in the overall aggregated score but also for each individual vocabulary. However, for ICD9-CM and ICD10-CM, Llama3-OpenBioLLM-70B outperformed random guessing. 

\subsection{Case Study: ICD9-CM Performance by Difficulty Level}
\label{sec:case_study_icd9}
Table \ref{tab:zs_fs_ICD9-CM_levels} illustrates a case study of the ICD9-CM vocabulary and the performance of the models across these levels (easy, medium, and hard). The objective of these experiments is to highlight the diverse range of difficulty levels represented by the questions within our proposed benchmark. As depicted in this table, the results of both GPT-4 and GPT-3.5 decrease as the level becomes more challenging. For easy questions, GPT-4 achieves very high accuracy scores, with 98\% for few-shot learning and 95.333\% for zero-shot learning. However, for medium questions, which are more difficult than the easy ones, GPT-4 achieves only 65.467\% for zero-shot and 72.4\% for few-shot learning. As for the hard level questions, the accuracy decreases dramatically, with only 47.467\% for zero-shot and 53.733\% for few-shot learning. This represents an absolute decrease of 44.267\% in few-shot accuracy between the easy and hard questions, and 47.866\% for zero-shot.

\begin{table}[ht]
\centering
\caption{Zero-shot and Few-shot learning results with 95\% confidence intervals, for ICD9-CM vocabulary.}
\scalebox{0.75}{
\begin{tabular}{|l|l|l|l||l|l|l|}
\hline
\multirow{2}{*}{Model} & \multicolumn{3}{c||}{Zero-shot} & \multicolumn{3}{c|}{Few-shot} \\
& Easy & Medium & Hard & Easy & Medium & Hard\\
\hline
BioMistral & $25.333 \pm 2.144$ & $24.933 \pm 4.091$ & $27.333 \pm 2.045$ & $25.6 \pm 2.759$ & $23.067 \pm 1.593$ & $26.8 \pm 2.357$ \\
BioBert & $23.333 \pm 2.237$ & $27.467 \pm 5.257$ & $26.133 \pm 3.343$ & $25.467 \pm 6.81$ & $25.067 \pm 1.833$ & $25.067 \pm 1.458$ \\
Llama3--OpenBioLLM - 70B &  $31.2 \pm 7.298$ & $27.467 \pm 9.174$ & $23.467 \pm 4.114$ & $36.4 \pm 0.454$  & $26.533 \pm 3.666$ & $27.2 \pm 1.571$ \\
BioMedGPT & $24.0 \pm 2.759$ & $25.067 \pm 2.579$ & $22.133 \pm 0.944$ & $24.533 \pm 5.412$ & $24.267 \pm 3.435$ & $24.267 \pm 2.046$ \\
Clinical-Longformer & $25.2 \pm 0.454$ & $24.933 \pm 3.404$ & $24.667 \pm 2.328$ & $28.4 \pm 1.636$ & $26.533 \pm 4.808$ & $26.4 \pm 1.2$ \\
GPT-3.5 & $64.0 \pm 0.454$ & $36.133 \pm 1.309$ & $40.533 \pm 1.593$ & $77.067 \pm 3.282$ & $40.267 \pm 3.282$ & $39.467 \pm 1.718$ \\
GPT-4 & $\textbf{95.333} \pm 2.328$ & $\textbf{65.467} \pm $2.283 & $\textbf{47.467} \pm 4.214$ & $\textbf{98.0} \pm 1.2$ & $\textbf{72.4} \pm 1.815$ & $\textbf{53.733} \pm 2.499$ \\
Gatortron & $24.667 \pm 4.763$ &  $26.933 \pm 4.806$ &  $26.133 \pm 2.498$ &  $25.6 \pm 2.078$ & $26.0 \pm 3.542$ & $24.667 \pm 3.464$ \\
MedMNX-7B & $23.6 \pm 5.499$ & $24.4 \pm 3.543$ & $22.8 \pm 4.374$ & $24.267 \pm 1.458$ & $25.067 \pm 1.593$ & $25.2 \pm 2.975$ \\
Meditron-70B & $27.733 \pm 4.655$ & $25.2 \pm 2.4$ & $26.533 \pm 3.694$ & $27.067 \pm 5.316$ & $24.133 \pm 4.634$ & $22.0 \pm 5.684$ \\
Meerkat & $25.2 \pm 2.079$ & $27.467 \pm 3.858$ & $25.333 \pm 2.498$ & $26.667 \pm 2.499$ & $27.067 \pm 0.693$ & $24.133 \pm 1.458$ \\
\hline
\end{tabular}}
\label{tab:zs_fs_ICD9-CM_levels}
\end{table}

\section{Discussion}

\begin{figure}[ht]
    \centering
    \begin{subfigure}[b]{0.3\linewidth}
        \includegraphics[width=\linewidth]{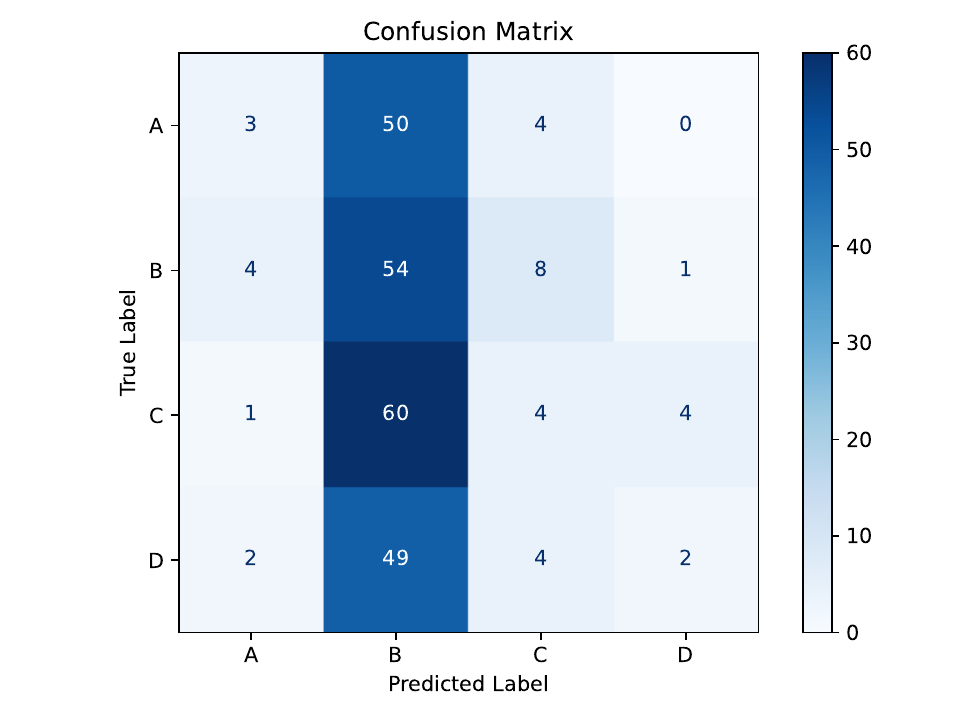}
        \caption{Meditron-70B error analysis.}
        \label{fig:error_analysis_icd9_easy_cllms_meditron}
    \end{subfigure}
    \begin{subfigure}[b]{0.3\linewidth}
        \includegraphics[width=\linewidth]{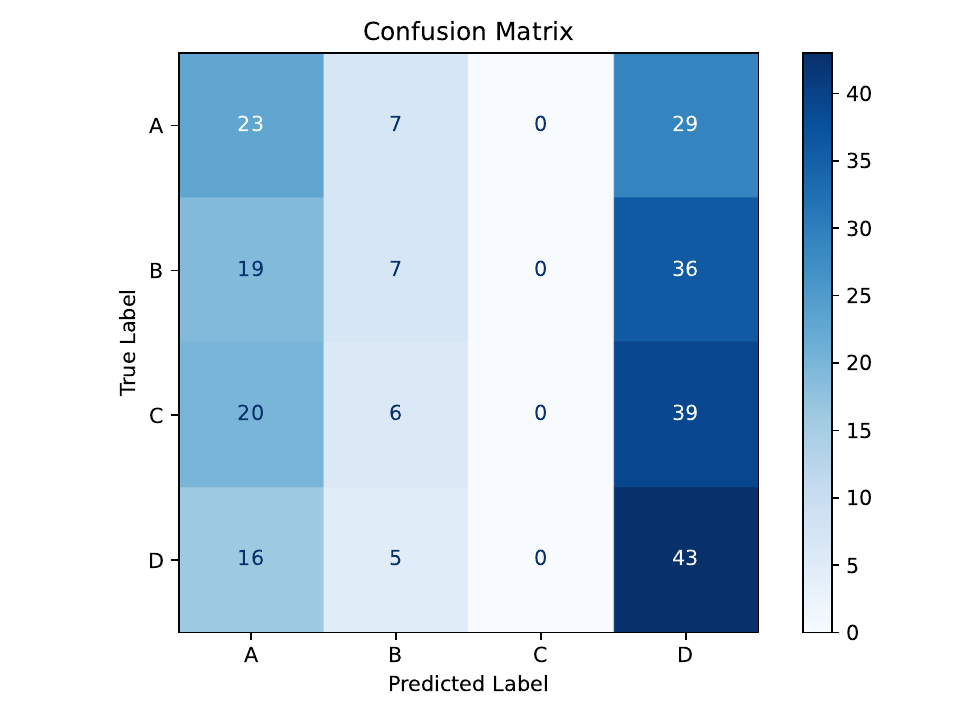}
        \caption{Meerkat-7B error analysis.}
        \label{fig:error_analysis_icd9_easy_cllms_meerkat}
    \end{subfigure}
    \begin{subfigure}[b]{0.3\linewidth}
        \includegraphics[width=\linewidth]{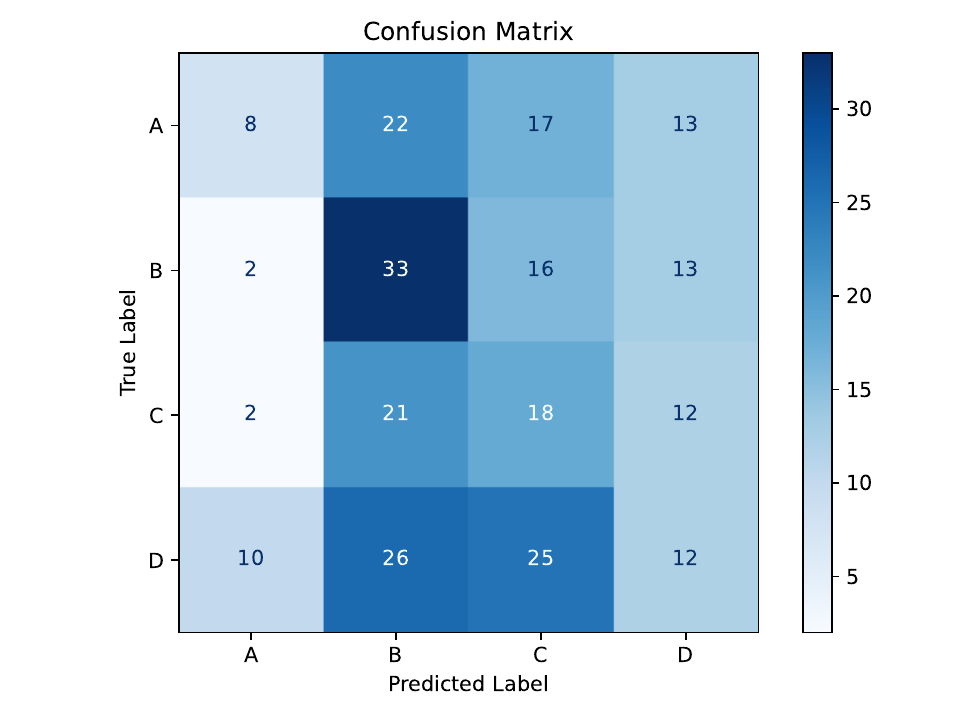}
        \caption{BioMistral-7B error analysis.}
        \label{fig:error_analysis_icd9_easy_cllms_biomistral}
    \end{subfigure}

    \caption{Error analysis (confusion matrix) of few-shot learning for ICD9-CM at the easy level, for three CLLMs that achieved random guessing performance on our benchmark.}
    \label{fig:error_analysis_icd9_easy_cllms}
\end{figure}

With the proposed benchmark in this work, we evaluate the understanding of medical concepts across LLMs, including those not pre-trained on medical data. Our experiments show that the current state-of-the-art CLLMs are not adequate in understanding the meaning of medical concepts such as diagnoses, procedures, and drug codes, and the differences between them. The performances of these CLLMs were close to random guessing. However, GPT-3.5 and GPT-4 achieved better accuracy for our benchmark, although the accuracy is still not high enough in some cases. For example, the few-shot learning accuracy of GPT-4 for ICD10-PROC and ICD9-PROC is only 44.489\% and 55.289\%, respectively. Additionally, we present a case study of the ICD9-CM vocabulary, which shows a significant performance difference between GPT-4 and GPT-3.5 models across different difficulty levels, demonstrating that our benchmark indeed contains questions of varying difficulty. 

In order to investigate why the CLLMs achieve a random guessing guess performance, we calculate the confusion matrix for three CLLMs (Meditron-70B, Meerkat-7B, and BioMistral-7B) for our case study vocabulary, ICD9-CM, under few-shot learning settings. We choose this level because it is the easiest level on the benchmark, and GPT-4 achieved very high accuracy, however, the CLLMs also achieved Random guessing performance on this level. Figure \ref{fig:error_analysis_icd9_easy_cllms} shows the error analysis using three confusion matrices for Meditron-70B, BioMistral-7B-DARE, and Meerkat-7B on the ICD9-CM easy level. Meditron-70B almost always predicts the label B, and Meerkat-7B almost always predicts A or D but never predicts C. BioMistral-7B does not always predict the same label but predicts diverse answers among the given options (A, B, C, D). However, the answers from this model are mostly incorrect since it also achieved Random guessing accuracy. Our careful dataset design, incorporating diverse label representation through random sampling in the 4-shot examples, ensures that we did not inadvertently influence the models (Meditron-70B and Meerkat-7B) to consistently predict the same label for this dataset. We encourage further studies to investigate this issue, focusing on how effectively existing CLLMs follow instructions, whether in zero-shot or few-shot learning scenarios.

% We provide a medical-codings Q\&A benchmark that can be used to evaluate and compare models...

\section{Limitations and Future Work}
% We didn't include all medical codes vobabulatires, for example LOINC which is vocabulary for identifying medical laboratory observations. In addition, our benchmark contains questions in a single form. 
% Future work: In this work we showed that CLLMs are achived random guess performance on MedConceptsQA benchmark. Future works should consider to imporove CLLMs to have an understanding of medical concepts, at least better than GPT-4 which his main focus is not medical in contrast to CLLMs. In addition, because our benchmark is competitive and provides an evaluation of whether a language model understands the meaning of medical concepts, we hope that new language models that also deal with medicine will test themselves on them and achieve better results than the current models.

\textbf{Limitations:} We did not include all medical code vocabularies, for example, LOINC, which is a vocabulary for identifying medical laboratory observations. Additionally, our benchmark contains questions in only a single form.

\textbf{Future Work:} In this work, we demonstrated that CLLMs achieved random guessing performance on the MedConceptsQA benchmark. Future research should focus on improving CLLMs to attain a better understanding of medical concepts, aiming for performance comparable to or better than GPT-4, whose primary focus is not medical, unlike CLLMs. Additionally, since our benchmark is competitive and evaluates a LLMs understanding of medical concepts, we encourage new CLLMs to include MedConceptsQA in their evaluation, with the aim of achieving superior results compared to current CLLMs.

\section{Conclusion}
In this study, we introduced MedConceptsQA, an open-source benchmark comprising over 800,000 questions spanning three difficulty levels. It was designed to evaluate the understanding and reasoning capabilities of LLMs on medical concepts across diagnoses, procedures, and drugs. Experimental results showed that state-of-the-art CLLMs, despite being pre-trained on medical data, achieved accuracy levels close to random guessing on this benchmark. However, general-purpose models (GPT-3.5 and GPT-4) outperformed CLLMs. Notably, GPT-4 exhibited the best performance, although its accuracy remained insufficient for certain datasets in our benchmark.

\section{Reproducibility}
The MedConceptsQA benchmark is available at \href{https://huggingface.co/datasets/ofir408/MedConceptsQA}{https://huggingface.co/datasets/ofir408/MedConceptsQA}. 

Our code for creating this benchmark, and the evaluation code are available at the following link: \href{https://github.com/nadavlab/MedConceptsQA}{https://github.com/nadavlab/MedConceptsQA}.

% references
\printbibliography

\section{Appendix}
% full results (zero & few shot, all models, all datasets, all levels, all experiments). 
% (set random_state in the evaluation?)
Table \ref{tab:zs_full_results} presents the comprehensive zero-shot learning results for all models, vocabularies, and difficulty levels (easy, medium, and hard). Table \ref{tab:fs_full_results} presents the few-shot learning results.

\begin{table}[h]
\centering
\resizebox{\textwidth}{!}{
\begin{tabular}{l|ccc|ccc|ccc|ccc|ccc}
\toprule
 & \multicolumn{3}{c|}{ICD9PROC} & \multicolumn{3}{c|}{ICD9CM} & \multicolumn{3}{c|}{ICD10PROC} & \multicolumn{3}{c|}{ICD10CM} & \multicolumn{3}{c}{ATC} \\
 & easy & medium & hard & easy & medium & hard & easy & medium & hard & easy & medium & hard & easy & medium & hard \\
\midrule
model & & & & & & & & & & & & & & & \\
BioMistral/BioMistral-7B-DARE & $22.4 \pm 2.08$ & $24.8 \pm 2.4$ & $25.2 \pm 3.54$ & $25.33 \pm 2.14$ & $24.93 \pm 4.09$ & $27.33 \pm 2.05$ & $22.13 \pm 3.52$ & $19.47 \pm 5.22$ & $26.0 \pm 4.08$ & $23.07 \pm 4.54$ & $24.27 \pm 2.88$ & $25.47 \pm 2.58$ & $22.8 \pm 2.97$ & $31.33 \pm 12.31$ & $24.0 \pm 3.6$ \\
HuggingFaceH4/zephyr-7b-beta & $23.6 \pm 2.4$ & $23.2 \pm 3.18$ & $26.67 \pm 0.52$ & $25.87 \pm 4.54$ & $27.07 \pm 2.33$ & $26.0 \pm 1.64$ & $25.2 \pm 3.54$ & $25.33 \pm 3.02$ & $26.0 \pm 6.68$ & $26.0 \pm 2.27$ & $26.13 \pm 4.36$ & $24.27 \pm 1.05$ & $26.4 \pm 3.63$ & $24.53 \pm 3.02$ & $26.8 \pm 3.95$ \\
PharMolix/BioMedGPT-LM-7B & $24.0 \pm 3.63$ & $23.73 \pm 6.02$ & $23.07 \pm 2.14$ & $24.0 \pm 2.76$ & $25.07 \pm 2.58$ & $22.13 \pm 0.94$ & $24.53 \pm 5.22$ & $29.47 \pm 9.70$ & $27.47 \pm 2.28$ & $22.8 \pm 2.72$ & $26.53 \pm 4.47$ & $25.6 \pm 2.4$ & $25.87 \pm 3.43$ & $21.73 \pm 9.99$ & $25.2 \pm 4.99$ \\
UFNLP/gatortron-large & $25.47 \pm 3.67$ & $27.73 \pm 1.39$ & $22.8 \pm 0.45$ & $24.67 \pm 4.76$ & $26.93 \pm 4.81$ & $26.13 \pm 2.50$ & $21.87 \pm 5.63$ & $24.53 \pm 2.33$ & $25.33 \pm 1.89$ & $25.87 \pm 2.33$ & $23.87 \pm 2.77$ & $26.53 \pm 3.28$ & $23.6 \pm 4.08$ & $19.6 \pm 3.42$ & $23.07 \pm 3.18$ \\
aaditya/Llama3-OpenBioLLM-70B & $25.87 \pm 4.89$ & $28.13 \pm 9.30$ & $24.53 \pm 4.12$ & $31.2 \pm 7.30$ & $27.47 \pm 9.18$ & $23.47 \pm 4.12$ & $25.47 \pm 3.18$ & $27.6 \pm 7.20$ & $22.8 \pm 7.87$ & $29.73 \pm 3.86$ & $26.27 \pm 4.54$ & $25.73 \pm 0.69$ & $27.33 \pm 4.59$ & $26.0 \pm 5.52$ & $22.4 \pm 1.2$ \\
dmis-lab/biobert-v1.1 & $25.2 \pm 4.54$ & $25.2 \pm 3.18$ & $27.73 \pm 3.43$ & $23.33 \pm 2.24$ & $27.47 \pm 5.26$ & $26.13 \pm 3.34$ & $26.27 \pm 2.50$ & $24.0 \pm 5.52$ & $24.93 \pm 4.54$ & $27.6 \pm 1.64$ & $28.13 \pm 5.06$ & $25.33 \pm 2.58$ & $26.27 \pm 4.12$ & $26.0 \pm 2.72$ & $28.67 \pm 2.92$ \\
dmis-lab/meerkat-7b-v1.0 & $24.67 \pm 3.67$ & $25.2 \pm 6.24$ & $22.4 \pm 4.74$ & $25.2 \pm 2.08$ & $27.47 \pm 3.86$ & $25.33 \pm 2.50$ & $23.73 \pm 1.39$ & $29.73 \pm 15.98$ & $27.6 \pm 2.53$ & $24.67 \pm 1.14$ & $28.67 \pm 1.39$ & $25.47 \pm 6.69$ & $25.2 \pm 2.97$ & $24.0 \pm 2.76$ & $25.87 \pm 1.83$ \\
epfl-llm/meditron-70b & $24.53 \pm 6.04$ & $26.67 \pm 5.06$ & $24.4 \pm 1.36$ & $27.73 \pm 4.65$ & $25.2 \pm 2.4$ & $26.53 \pm 3.69$ & $24.93 \pm 1.31$ & $27.87 \pm 2.10$ & $22.53 \pm 0.52$ & $25.87 \pm 1.05$ & $23.07 \pm 0.52$ & $25.6 \pm 0.79$ & $27.87 \pm 0.52$ & $23.73 \pm 2.62$ & $23.87 \pm 6.81$ \\
epfl-llm/meditron-7b & $25.33 \pm 3.86$ & $25.87 \pm 2.05$ & $27.2 \pm 2.83$ & $26.53 \pm 0.69$ & $24.93 \pm 2.28$ & $25.07 \pm 2.50$ & $23.07 \pm 4.98$ & $25.47 \pm 4.07$ & $24.8 \pm 1.36$ & $25.47 \pm 1.72$ & $24.67 \pm 9.13$ & $28.27 \pm 3.80$ & $26.0 \pm 2.4$ & $27.73 \pm 5.16$ & $25.87 \pm 3.28$ \\
gpt-3.5-turbo & $31.2 \pm 2.76$ & $33.33 \pm 1.14$ & $29.07 \pm 3.67$ & $64.0 \pm 0.45$ & $36.13 \pm 1.31$ & $40.53 \pm 1.59$ & $24.0 \pm 2.97$ & $35.87 \pm 3.67$ & $28.53 \pm 2.05$ & $68.27 \pm 3.02$ & $37.6 \pm 4.80$ & $37.07 \pm 2.05$ & $35.33 \pm 2.73$ & $25.33 \pm 0.94$ & $29.6 \pm 2.83$ \\
gpt-4-0125-preview & $54.0 \pm 6.28$ & $40.13 \pm 4.09$ & $35.73 \pm 0.94$ & $95.33 \pm 2.33$ & $65.47 \pm 2.28$ & $47.47 \pm 4.21$ & $30.0 \pm 3.14$ & $42.27 \pm 3.43$ & $34.93 \pm 1.14$ & $94.27 \pm 2.14$ & $54.53 \pm 4.54$ & $56.4 \pm 2.4$ & $57.87 \pm 3.02$ & $45.33 \pm 4.98$ & $33.6 \pm 2.08$ \\
johnsnowlabs/JSL-MedMNX-7B & $22.67 \pm 3.22$ & $23.47 \pm 1.39$ & $22.53 \pm 2.58$ & $23.6 \pm 5.50$ & $24.4 \pm 3.54$ & $22.8 \pm 4.37$ & $27.07 \pm 1.72$ & $23.6 \pm 3.54$ & $26.27 \pm 2.10$ & $25.47 \pm 3.02$ & $24.67 \pm 4.07$ & $24.4 \pm 1.2$ & $26.13 \pm 1.46$ & $24.93 \pm 4.29$ & $24.4 \pm 5.79$ \\
meta-llama/Meta-Llama-3-8B-Instruct & $24.0 \pm 1.57$ & $28.27 \pm 6.93$ & $27.73 \pm 5.83$ & $26.67 \pm 8.91$ & $25.87 \pm 3.46$ & $26.67 \pm 3.64$ & $22.13 \pm 2.92$ & $32.67 \pm 5.22$ & $22.0 \pm 9.47$ & $27.6 \pm 13.75$ & $22.67 \pm 6.42$ & $24.8 \pm 9.81$ & $26.67 \pm 5.67$ & $25.2 \pm 5.52$ & $24.67 \pm 3.86$ \\
yikuan8/Clinical-Longformer & $26.4 \pm 2.4$ & $27.2 \pm 1.2$ & $24.0 \pm 5.11$ & $25.2 \pm 0.45$ & $24.93 \pm 3.40$ & $24.67 \pm 2.33$ & $24.4 \pm 3.6$ & $22.93 \pm 3.78$ & $24.27 \pm 2.33$ & $28.4 \pm 2.53$ & $24.0 \pm 2.08$ & $24.0 \pm 1.36$ & $25.73 \pm 1.59$ & $23.2 \pm 0.91$ & $26.27 \pm 3.02$ \\
\bottomrule
\end{tabular}
}
\caption{Zero-shot learning full results for all the models, vocabularies, and levels.}
\label{tab:zs_full_results}
\end{table}

\begin{table}
\centering
\resizebox{\textwidth}{!}{

\begin{tabular}{l|ccc|ccc|ccc|ccc|ccc}
\hline
& \multicolumn{3}{c|}{ICD9PROC} & \multicolumn{3}{c|}{ICD9CM} & \multicolumn{3}{c|}{ICD10PROC} & \multicolumn{3}{c|}{ICD10CM} & \multicolumn{3}{c}{ATC} \\
& easy & medium & hard & easy & medium & hard & easy & medium & hard & easy & medium & hard & easy & medium & hard \\
\hline
model & & & & & & & & & & & & & & & \\
BioMistral/BioMistral-7B-DARE & 29.467 $\pm$ 2.045 & 23.867 $\pm$ 3.694 & 25.6 $\pm$ 2.759 & 25.6 $\pm$ 2.759 & 23.067 $\pm$ 1.593 & 26.8 $\pm$ 2.356 & 26.8 $\pm$ 2.078 & 22.0 $\pm$ 2.078 & 21.067 $\pm$ 3.91 & 24.0 $\pm$ 3.6 & 26.0 $\pm$ 2.356 & 22.8 $\pm$ 3.142 & 24.8 $\pm$ 0.907 & 27.867 $\pm$ 5.707 & 26.133 $\pm$ 1.141 \\
HuggingFaceH4/zephyr-7b-beta & 25.867 $\pm$ 3.02 & 23.733 $\pm$ 3.434 & 26.8 $\pm$ 0.454 & 24.133 $\pm$ 1.047 & 26.933 $\pm$ 6.257 & 25.6 $\pm$ 1.977 & 24.8 $\pm$ 2.759 & 23.2 $\pm$ 0.785 & 25.333 $\pm$ 2.045 & 23.333 $\pm$ 0.944 & 27.333 $\pm$ 2.498 & 26.533 $\pm$ 2.143 & 23.6 $\pm$ 1.2 & 24.933 $\pm$ 2.095 & 23.733 $\pm$ 1.141 \\

PharMolix/BioMedGPT-LM-7B & 23.467 $\pm$ 2.916 & 29.733 $\pm$ 7.539 & 24.667 $\pm$ 1.833 & 24.533 $\pm$ 5.41 & 24.267 $\pm$ 3.434 & 24.267 $\pm$ 2.045 & 24.4 $\pm$ 1.2 & 22.533 $\pm$ 3.962 & 24.667 $\pm$ 5.67 & 24.533 $\pm$ 1.593 & 22.4 $\pm$ 2.4 & 23.6 $\pm$ 1.635 & 26.267 $\pm$ 3.281 & 26.133 $\pm$ 4.543 & 28.4 $\pm$ 2.974 \\

UFNLP/gatortron-large & 25.2 $\pm$ 3.6 & 23.2 $\pm$ 2.525 & 24.533 $\pm$ 4.358 & 25.6 $\pm$ 2.078 & 26.0 $\pm$ 3.542 & 24.667 $\pm$ 3.464 & 25.067 $\pm$ 1.888 & 25.2 $\pm$ 3.175 & 23.867 $\pm$ 2.951 & 24.8 $\pm$ 1.635 & 24.0 $\pm$ 3.542 & 24.933 $\pm$ 0.693 & 26.667 $\pm$ 3.217 & 23.333 $\pm$ 2.283 & 25.867 $\pm$ 0.944 \\

aaditya/Llama3-OpenBioLLM-70B & 29.467 $\pm$ 3.217 & 25.333 $\pm$ 0.693 & 22.533 $\pm$ 1.141 & 36.4 $\pm$ 0.454 & 26.533 $\pm$ 3.666 & 27.2 $\pm$ 1.571 & 24.667 $\pm$ 2.951 & 26.267 $\pm$ 1.309 & 28.667 $\pm$ 2.951 & 31.867 $\pm$ 3.02 & 26.8 $\pm$ 3.6 & 24.8 $\pm$ 1.977 & 27.067 $\pm$ 3.343 & 23.067 $\pm$ 3.694 & 22.4 $\pm$ 0.785 \\

dmis-lab/biobert-v1.1 & 28.533 $\pm$ 3.343 & 23.867 $\pm$ 1.141 & 23.867 $\pm$ 2.327 & 25.467 $\pm$ 6.808 & 25.067 $\pm$ 1.833 & 25.067 $\pm$ 1.458 & 26.667 $\pm$ 1.385 & 25.6 $\pm$ 3.424 & 26.267 $\pm$ 1.047 & 24.8 $\pm$ 2.4 & 27.333 $\pm$ 3.087 & 27.333 $\pm$ 3.91 & 26.0 $\pm$ 4.799 & 22.4 $\pm$ 1.977 & 23.6 $\pm$ 0.785 \\

dmis-lab/meerkat-7b-v1.0 & 26.4 $\pm$ 5.948 & 24.533 $\pm$ 1.717 & 24.933 $\pm$ 3.694 & 26.667 $\pm$ 2.498 & 27.067 $\pm$ 0.693 & 24.133 $\pm$ 1.458 & 24.267 $\pm$ 1.141 & 27.6 $\pm$ 7.71 & 23.867 $\pm$ 2.579 & 26.933 $\pm$ 2.498 & 22.133 $\pm$ 3.666 & 23.733 $\pm$ 4.065 & 23.467 $\pm$ 2.045 & 23.467 $\pm$ 2.143 & 24.933 $\pm$ 1.309 \\

epfl-llm/meditron-70b & 25.2 $\pm$ 2.078 & 25.867 $\pm$ 0.944 & 26.533 $\pm$ 3.281 & 27.067 $\pm$ 5.315 & 24.133 $\pm$ 4.632 & 22.0 $\pm$ 5.682 & 27.067 $\pm$ 2.045 & 24.8 $\pm$ 3.927 & 25.333 $\pm$ 1.888 & 25.867 $\pm$ 3.962 & 26.133 $\pm$ 7.045 & 25.867 $\pm$ 4.72 & 24.533 $\pm$ 2.143 & 24.4 $\pm$ 2.974 & 24.133 $\pm$ 1.833 \\

epfl-llm/meditron-7b & 24.4 $\pm$ 1.361 & 26.133 $\pm$ 3.803 & 25.6 $\pm$ 0.907 & 24.4 $\pm$ 2.721 & 22.8 $\pm$ 4.031 & 24.0 $\pm$ 1.2 & 25.467 $\pm$ 2.771 & 22.267 $\pm$ 9.991 & 23.733 $\pm$ 4.214 & 21.867 $\pm$ 4.632 & 24.133 $\pm$ 3.637 & 22.533 $\pm$ 3.02 & 24.667 $\pm$ 2.657 & 21.067 $\pm$ 5.157 & 23.733 $\pm$ 2.327 \\

gpt-3.5-turbo & 31.733 $\pm$ 6.09 & 31.067 $\pm$ 2.045 & 30.8 $\pm$ 0.907 & 77.067 $\pm$ 3.281 & 40.267 $\pm$ 3.281 & 39.467 $\pm$ 1.717 & 27.333 $\pm$ 0.262 & 28.533 $\pm$ 2.771 & 29.067 $\pm$ 2.771 & 76.0 $\pm$ 4.326 & 39.6 $\pm$ 1.635 & 39.067 $\pm$ 2.657 & 67.067 $\pm$ 0.944 & 39.2 $\pm$ 1.571 & 25.867 $\pm$ 2.951 \\

gpt-4-0125-preview & 80.0 $\pm$ 4.467 & 50.8 $\pm$ 1.814 & 35.067 $\pm$ 3.434 & 98.0 $\pm$ 1.2 & 72.4 $\pm$ 1.814 & 53.733 $\pm$ 2.498 & 46.8 $\pm$ 3.424 & 52.267 $\pm$ 1.458 & 34.4 $\pm$ 1.571 & 98.133 $\pm$ 0.944 & 63.867 $\pm$ 2.498 & 57.067 $\pm$ 2.88 & 92.4 $\pm$ 0.785 & 59.467 $\pm$ 4.543 & 34.267 $\pm$ 1.458 \\

johnsnowlabs/JSL-MedMNX-7B & 25.067 $\pm$ 0.262 & 25.6 $\pm$ 2.356 & 28.133 $\pm$ 3.523 & 24.267 $\pm$ 1.458 & 25.067 $\pm$ 1.593 & 25.2 $\pm$ 2.974 & 24.133 $\pm$ 3.857 & 27.6 $\pm$ 1.2 & 24.667 $\pm$ 3.185 & 26.0 $\pm$ 3.175 & 27.333 $\pm$ 4.065 & 25.6 $\pm$ 2.356 & 22.4 $\pm$ 2.078 & 27.867 $\pm$ 4.214 & 25.467 $\pm$ 1.141 \\

meta-llama/Meta-Llama-3-8B-Instruct & 27.6 $\pm$ 3.175 & 24.933 $\pm$ 5.41 & 25.6 $\pm$ 0.785 & 27.067 $\pm$ 1.833 & 24.267 $\pm$ 1.593 & 26.133 $\pm$ 1.141 & 24.8 $\pm$ 0.907 & 28.933 $\pm$ 4.995 & 26.133 $\pm$ 2.237 & 23.333 $\pm$ 0.944 & 21.867 $\pm$ 0.944 & 26.267 $\pm$ 4.654 & 26.133 $\pm$ 1.458 & 26.0 $\pm$ 8.276 & 25.733 $\pm$ 2.237 \\

yikuan8/Clinical-Longformer & 26.0 $\pm$ 4.031 & 24.8 $\pm$ 2.525 & 26.4 $\pm$ 1.977 & 28.4 $\pm$ 1.635 & 26.533 $\pm$ 4.807 & 26.4 $\pm$ 1.2 & 23.867 $\pm$ 3.857 & 24.933 $\pm$ 0.944 & 25.2 $\pm$ 2.078 & 24.533 $\pm$ 2.498 & 26.4 $\pm$ 7.71 & 25.733 $\pm$ 5.448 & 24.267 $\pm$ 2.095 & 24.4 $\pm$ 5.517 & 25.333 $\pm$ 6.09 \\
\hline
\end{tabular}
}
\caption{Few-shot learning full results for all the models, vocabularies, and levels.}
\label{tab:fs_full_results}
\end{table}

%%%%%%%%%%%%%%%%%%%%%%%%%%%%%%%%%%%%%%%%%%%%%%%%%%%%%%%%%%%%

\end{document}